\documentclass[accepted]{uai2022}

\usepackage[american]{babel}

\usepackage{times} %
\usepackage{helvet} %
\usepackage{courier} %
\usepackage{graphicx} %
\usepackage{natbib}  %
\usepackage{caption}  %
\DeclareCaptionStyle{ruled}%
 {labelfont=normalfont,labelsep=colon,strut=off}

\usepackage{soul}
\usepackage[utf8]{inputenc}
\usepackage{graphicx}
\usepackage{amsmath}
\usepackage{amsthm}
\usepackage{mathrsfs}
\usepackage{cleveref}
\usepackage{booktabs}
\usepackage{algorithm}
\usepackage{algorithmic}
\usepackage{bm}
\usepackage{bbm}
\usepackage{subfig}
\usepackage{amssymb}
\usepackage{enumitem}
\usepackage{tikz}
\usepackage{soul}
\usepackage{float}
\newcommand{\indep}{\rotatebox[origin=c]{90}{$\models$}}

\theoremstyle{dfn}
\newtheorem{dfn}{Definition}

\theoremstyle{proposition}

\newtheorem{theorem}{Theorem}
\newtheorem{corrollary}{Corollary}

\newtheorem{assumption}{A}

\newcommand{\commentout}[1]{%
}

\newcommand{\Cov}{\mathrm{Cov}}

\newcommand{\ba}{Barab\'asi-Albert}
\newcommand{\er}{Erd\H{o}s-R\'{e}nyi}

\newcommand{\mcit}{NIRD}

\newcommand{\rel}[1]{\sigma_{#1}(v_i)}
\newcommand{\relv}[1]{\sigma_{#1}^{v_i}}

\title{Non-Parametric Inference of Relational Dependence}

\author[1]{Ragib Ahsan}
\author[1]{Zahra Fatemi}
\author[2]{David Arbour}
\author[1]{Elena Zheleva}
\affil[1]{%
    Department of Computer Science\\
    University of Illinois at Chicago\\
    Chicago, IL, USA
}
\affil[2]{%
    Adobe Research, USA
}

\begin{document}

\maketitle

\begin{abstract}
Independence testing plays a central role in statistical and causal inference from observational data. Standard independence tests assume that the data samples are independent and identically distributed (i.i.d.) but that assumption is violated in many real-world datasets and applications centered on relational systems. 
This work examines the problem of estimating independence in data drawn from relational systems by defining sufficient representations for the sets of observations influencing individual instances.
Specifically, we define marginal and conditional independence tests for relational data by considering the kernel mean embedding as a flexible aggregation function for relational variables. 
We propose a consistent, non-parametric, scalable kernel test to operationalize the relational independence test for non-i.i.d. observational data under a set of structural assumptions.
We empirically evaluate our proposed method on a variety of synthetic and semi-synthetic networks and demonstrate its effectiveness compared to state-of-the-art kernel-based independence tests. 
\end{abstract}

\section{Introduction}
\label{sec:intro}

Measuring dependence is a fundamental task in statistics. However, most existing independence tests assume that the observed data is independent and identically distributed (i.i.d.). This assumption makes them unsuitable for capturing statistical dependencies in real-world relational systems, from social networks to protein-protein interactions, in which data instances depend on each other. %

Relational dependence refers to a statistical dependence, marginal or conditional, between random variables in which at least one of the variables is %
a relational variable~\citep{lee-uai17}. A relational variable is a set %
of random variables %
that belong to instances related %
to an instance of interest, such as friends of a person or proteins interacting with a target protein. Relational dependence testing is central to social influence studies and causal discovery in relational systems, yet there is no standard statistical tool for inferring different forms of dependence from observational relational data. In this paper, we present a practical tool for determining marginal and conditional independence between
relational variables with consistency guarantees.

To understand the challenge of estimating relational dependence, let's consider the following example:

\noindent

\textbf{Example 1.} \textit{Sally doesn't smoke but she has some friends who smoke. Sally starts to smoke over time. Are the smoking habits of Sally's friends and Sally's decision to take up smoking independent?}

Part of the problem of detecting this dependence is that we need to know a priori the exact mechanism of dependence. Is it because all of Sally's friends share a certain behavior? Or is there a minimum threshold of friends necessary (e.g., half of her friends smoke) to activate the dependence? 
The nature of the relationships among her friends might play a role as well. Without prior knowledge of this mechanism of dependence, any existing statistical test is likely to fail. As such, the focus of our work is to develop a flexible non-parametric test that can capture multiple forms of dependence. Note that for the scope of this work, we do not focus on the causal aspects of this question (e.g., social influence) but the test that we propose is applicable to causal discovery as well. 

Recent studies have proposed non-parametric independence tests designed for non-i.i.d data.~\citep{flaxman-tist16,lee-uai17}. %
\citet{flaxman-tist16} develop a test between propositional variables accounting for latent homophily in a grid network but do not consider relational variables.
\citet{lee-uai17} develop a test for conditional independence (KRCIT) between relational variables which is the current state-of-the-art test for relational dependence. They operationalize the test by flattening the relational data into a single, propositional table.
However, the current state-of-the-art test has three key limitations. First, it requires practitioners to make explicit assumptions about the data generating processes and to specify an aggregation function over the relational variable a priori. Second, existing tests rely on propositionalization, which refers to the process of projecting connected data to a single, propositional table, which raises statistical concerns~\cite{maier-win13}. Third, it is computationally expensive and inapplicable to large relational datasets.

In this work, we focus on developing a general definition of relational dependence and a statistical test, NIRD (Non-parametric inference of relational dependence), 
that is able to capture a family of aggregate functions for characterizing relational dependence.
The contributions of our work are in providing 1) complete definitions of marginal and conditional relational dependence which extend the classical definition of ~\citet{daudin-biometrika80}, 2) independence tests with consistency guarantees, 3) the explicit representation of neighboring sets through kernel mean embeddings, and 4) a scalable test through random Fourier feature formulation which makes the test practical for real-world applications.
We compare our proposed test to KRCIT on a variety of synthetic networks and simulate several social network characteristics, such as structure, density, and size. We demonstrate the applicability of our method for detecting peer influence using both semi-synthetic and real-world social networks.

\section{Related Work}

There are several independence tests from observational data. Partial correlation is used for the independence test on Gaussian variables with linear dependence~\citep{baba-anjs04}. A more advanced test, 
the Hilbert-Schmidt independence criterion (HSIC)~\citep{gretton-icalt05} is a statistical test that has been extended to marginal independence testing for structured data~\citep{zhang-nips09} and random processes~\citep{chwialkowski-nips14}. 
\citet{flaxman-tist16} utilize HSIC to develop marginal and conditional independence tests for propositional variables in the presence of a latent confounder in a single-entity network with an additive noise generating function. %
Relational dependence exhibits itself through a pairwise edge in the network which is a noisy surrogate of latent homophily. %
~\citet{lee-uai17} extend these tests %
to explicitly include relational variables. They operationalize the tests by considering pairwise dependence between each member of the relational variable set in the flattened data.
However, existing tests require prior domain knowledge in order to  detect complex dependencies between relational variables such as entropy, variance or reaching a threshold of peers’ actions by requiring the specification of aggregation functions.
We lessen this requirement by employing kernel mean embeddings, which provide a non-parametric aggregation capable of capturing a large number of aspects of smooth functions over distributions belonging to the exponential family.

Relational independence testing is central to causal discovery and structure learning of relational models. 
However, existing relational causal discovery algorithms either rely on the existence of a relational independence oracle ~\citep{maier-aaai10,maier-uai13,lee-aaai16} or use conditional independence tests developed for i.i.d. data~\citep{lee-uai19}, making them less practical for real-world applications. %
The relational dependence tests in this work together with existing tests for detecting causal direction of relational dependence~\citep{arbour2016inferring} can make relational causal discovery achievable for real-world datasets with unknown dependence functions. 

One of the applications of relational dependence testing is in characterizing social influence 
in observational relational data. Existing social influence studies measure influence through predictive models~\citep{christakis-nejm07,bakshy-wsdm11}, or randomized controlled trials~\citep{muchnik-science13,su-icwsm20}. A 
standard statistical tool to infer different forms of social influence from observational relational data does not exist. Our work can help develop such tools by treating social influence detection as an instance of a relational independence test.

\section{Problem Definition}
\label{sec:problem}
The central task of this paper is to develop a statistical test for determining marginal and conditional independence between relational variables. Before formalizing the problem, we introduce the necessary notations and definitions. We denote random variables and their realizations with uppercase and lowercase letters, respectively, and bold to denote sets. We use an Entity-Relationship model~\citep{heckerman-isrl07} to describe relational data following previous work~\citep{maier-aaai10,lee-uai17}. A relational schema $\mathcal{S} = \langle \bm{\mathcal{E}}, \bm{\mathcal{R}}, \bm{\mathcal{A}} \rangle$ represents a relational domain where $\bm{\mathcal{E}}$, $\bm{\mathcal{R}}$ and $\bm{\mathcal{A}}$ refer to the set of entity, relationship and attribute classes respectively. We consider an undirected graph $G$ %
to be the instantiation of the relational schema $\mathcal{S}$ where nodes and edges refer to the entity and relationship instances respectively. A given schema can entail numerous possible instantiations.  
For ease of exposition, we focus on a schema with a single entity type (e.g., Person) and a single relationship type (e.g., Friends) but discuss the extensions necessary to address multi-relational data in the Appendix. %
We refer to the set of vertices, edges and adjacency matrix of $G$ as $\bm{V}$, $\bm{E}$ and $A$ respectively. For example, nodes $v_1=Ana$ and $v_2=Bob$ with undirected edge between them can come from the $Person - Friends - Person$ relations. %
The attribute class $\bm{\mathcal{A}}$ represents the set of possible attributes for the specified entity of the given graph $G$. Let $v_i.X$ and $v_i.x$ refer to the attribute $X \in \bm{\mathcal{A}}$ and its realization respectively for node $v_i \in \bm{V}$ corresponding to instance $i \in \bm{\mathcal{E}} \cup \bm{\mathcal{R}}$. Here, $v_i.X$ is considered to be a propositional random variable. Following prior work we define a relational variable as a set of propositional random variables~\citep{maier-uai13,lee-uai17}.

\begin{dfn}[Relational Variable]
\label{dfn:relvar}
Given a relational schema $\mathcal{S} = \langle \bm{\mathcal{E}}, \bm{\mathcal{R}}, \bm{\mathcal{A}} \rangle$, its instantiation $G$ and a path predicate $\rho$, a relational variable $\sigma(v_i, \bm{X}, G, \rho)$ is the set of attributes $v_j.\bm{X}$ selected by $\rho$ of nodes $v_j \in \mathbf{V}$ reachable from $v_i \in \mathbf{V}$  such that $\bm{X} \subseteq \bm{\mathcal{A}}$, where the path predicate $\rho$ is a function given by: $\rho(v_i, G) : \mathbf{V} \mapsto \mathcal{P}(\mathbf{V})$.

\end{dfn}

Here, $\mathcal{P}(\mathbf{V})$ refers to the power set of $\mathbf{V}$. An example path predicate is $\rho(v_i, G) = \{v_j | v_j \in \hat{\mathcal{N}}(v_i)\}$ where $\hat{\mathcal{N}}(v_i)$ refers to the direct neighbors of $v_i$ in $G$ and the corresponding relational variable $\sigma(v_i, \bm{X}, G, \rho)$ refers to the set of attributes of the neighboring nodes. For simplicity and assuming only direct neighbors throughout the paper, we denote the relational variable corresponding to attribute $X$ by $\rel{X}$ and its value by $\relv{x}$. Note that $\rel{X}$ can represent a \textit{propositional} variable as a special case. For example, $\rel{X} = \{v_i.X\}$ refers to the $X$ attribute of $v_i$. 
We also make the following assumptions:
\begin{assumption}
\label{assump:connected}
Each node $v\in V$ has degree of at least 1.
\end{assumption}
\begin{assumption}
\label{assump:binary}
The adjacency matrix of $G$ is symmetric with edge weights bounded by some real constant. 
\end{assumption}

\begin{assumption}
Dependence between two instances $i$ and $j$ implies the existence of a path in the graph between $v_i$ and $v_j$.
\end{assumption}

{Relational dependence refers to a statistical dependence, either marginal or conditional, between two variables where at least one of the variables is relational. The goal of a relational dependence test is to determine whether to reject the null hypothesis of independence between these variables or not. The representation of relational data for such a test is non-trivial because data instances are not i.i.d. A common practice to deal with relational data is \textit{propositionalization} ~\citep{kramer-rdm01}, %
which refers to the process of projecting a set of connected data samples down to a single, propositional table. %
In the context of relational dependence testing, flattening has three main deficiencies. First, the entities in the flattened data are not i.i.d. Second, choosing the appropriate aggregation function is non-trivial as discussed in section \ref{sec:intro}.
Failing to appropriately define the aggregate in this case could lead to increased type I errors in marginal tests, and both type I and II errors for conditional tests. Third, flattening raises statistical concerns for relational causal discovery, one of the application areas of relational conditional independence tests, by violating the causal Markov condition~\citep{maier-win13}.
~\citet{lee-uai17} address the first deficiency by proposing a solution framework based on graph kernels using an existing i.i.d. kernel-based CI test method. However, their approach does not directly address the other two concerns.
}

Let's look at the problem with a concrete %
example. We consider an entity Person which exhibits attributes such as 
smoking status before ($S0$) and after ($S1$) a given time threshold $t$ and $G$ represents the network of social ties. Detecting the dependence of peers on a person's smoking behavior can be formalized as an independence test. For example, detecting whether one's smoking behaviour is marginally independent of one's direct friends' smoking behaviour could be carried out by a marginal test of $v_i.S1 \indep \rel{S0}$. 
Similarly, a conditional test of $v_i.S1 \indep \rel{S0} | v_i.S0$ should detect whether one's current smoking behavior is independent of friends' old smoking behavior given one's old smoking behavior.

In this work, we propose a relational dependence test which captures complex dependencies between relational variables without relying on flattening or explicit aggregate representations. We extend the definition of conditional independence for non-parametric functions by \citet{daudin-biometrika80} %
and propose the following definitions of marginal and \textit{relational} conditional independence:

\begin{dfn}[Relational Marginal Independence]
\label{dfn:reldep_marg}
Two relational variables, $\rel{X}$ and $\rel{Y}$ are said to be marginally independent of each other if and only if,
$\mathbb{E}\left[ g_X(\rel{X})g_Y(\rel{Y}) \right]  = \mathbb{E}\left[g_X(\rel{X}) \right] \mathbb{E}\left[g_Y(\rel{Y}) \right]$ for any smooth square measurable functions $g_X(\cdot), g_Y(\cdot)$.
\end{dfn}
\begin{dfn}[Relational Conditional Independence]
\label{dfn:reldep_cond}
Two relational variables, $\rel{X}$ and $\rel{Y}$ are said to be independent of each other given a third, $\rel{Z}$ if and only if,
$\mathbb{E}\left[g_X(\rel{X})g_Y(\rel{Y}) | g_Z(\rel{Z}) \right]  = \mathbb{E}\left[g_X(\rel{X}) | g_Z(\rel{Z}) \right] \mathbb{E}\left[g_Y(\rel{Y}) | g_Z(\rel{Z}) \right]$ for any smooth square measurable functions $g_X(\cdot), g_Y(\cdot), g_Z(\cdot)$.
\end{dfn}

Here, $g_X(\cdot), g_Y(\cdot), g_Z(\cdot)$ are \textit{aggregate} functions that map $\sigma$ to a real-valued vector. They could be \textit{sum}, \textit{mean} or any other complex non-linear function. 
The rejection of the null hypothesis of marginal independence would mean that the variables are possibly dependent, either due to a directed path between them or due to a direct, causal relationship, or the presence of a confounding relationship. For a relational conditional independence (RCI) test, the rejection of the null hypothesis would imply that the two variables are not independent given the conditioning set. Note that because we are considering the dependence between sets of relational variables and their propositional counterparts we circumvent the three problems with flattening described earlier.

\section{Relational Independence Tests}
\label{sec:solution}

In this section, we discuss the components which operationalize the definition of relational dependence into an empirical test. We first describe a non-parametric relational aggregate formed by local kernel means. 
Then we formulate marginal and conditional independence tests using the kernel mean embedding.
Then, we discuss the theoretical boundaries for the consistency of the proposed test. Finally, we introduce techniques for large-scale approximation of the proposed relational kernels that can speed up the independence test significantly.

\subsection{Non-parametric Aggregate Representations}
One of the central problems in estimating dependence in relational settings is defining a sufficient representation for the sets of observations for individual instances of a relational variable. 
Prior work \citep{maier-uai13,arbour-kdd16,lee-uai17} considered aggregation functions, usually one, which are specified \textit{apriori} by the practitioner.
However, in many scenarios, it is unreasonable to expect practitioners to reason over a very complex joint distribution or to know the exact parametric form of dependence. %
For example, the possible aggregation in effect for the spread of obesity in social networks~\citep{christakis-nejm07} can be different from people’s influence on the Twitter platform~\citep{bakshy-wsdm11}. A generalized definition and associated operationalization of relational dependence can help the practitioner by directly measuring dependence without prior domain knowledge about aggregations on the given relational system. %

The distance between the embedding of joint distribution and embedding of product of marginals can be used to infer independence according to definition \ref{dfn:reldep_marg}, while avoiding explicit density estimation as an intermediate step. Note that the aggregate functions are represented implicitly through the kernel mean embedding (KME)~\citep{smola2007hilbert,muandet-ftml17}. %
An appealing property of the kernel mean embedding is that if the kernel is universal\mark{} then the kernel mean uniquely represents all moments for any member of the exponential family~\citep{smola2007hilbert}
\footnote{We refer readers to \citet{szabo2017characteristic} for conditions for a kernel to be universal. Many popular kernels such as the RBF kernel are universal}.

Adopting the kernel mean as an aggregation function removes the burden of reasoning over parametric families and predefined aggregates. Specifically, the kernel mean embedding considers the mean of a variable after applying a projection $\phi(\cdot)$ into some RKHS, $\mathbf{\mu} = \int \phi(x) p(x)dx$, with the corresponding empirical estimate of $\hat{\mathbf{\mu}} = \frac{1}{N}\sum_i^N \phi(x_i)$ where $N$ is the number of observations and $x_1,\dots, x_N$ are observations from a random variable $X$~\citep{smola2007hilbert}.

We present the practical implementation of the kernel mean as a relational aggregate. 
For a given node $v_i$, we define the kernel mean aggregate of its neighbors with respect to the attribute $X$ as 
$
   \mu(v_i) = \frac{1}{\deg{(v_i)}}\sum_{m \in \hat{\mathcal{N}}(v_i)} \phi(m.x)
$ 

where $\hat{\mathcal{N}}(\cdot)$ refers to a path predicate which is restricted to immediate neighbors for ease of exposition. Because $\phi$ may map to an infinite dimension, it is impractical to explicitly represent this quantity. 
Fortunately, because our statistics of interest are concerned with the covariance, the kernel trick, i.e. considering the inner product rather than the feature representations directly, can be employed. 
Specifically, the inner product between relational kernel mean is given as
\begin{align*}
    \label{eq:innerprodkernmean}
    &\langle \mu(v_i), \mu(v_j) \rangle = \frac{\sum_{m \in \hat{\mathcal{N}}(v_i)}\sum_{p \in \hat{\mathcal{N}}(v_j)} k(m.x, p.x)}{\textrm{deg}(v_i)\textrm{deg}(v_j)},
\end{align*}
which can be written for an entire sample in terms of a matrix product between the network adjacency matrix, $A$, the inverse degree matrix $D^{-1}$ where $D_{i,i} = \frac{1}{\text{deg}(v_i)}$, and the kernel matrix $K_X$, by observing
$
    (D^{-1}A\phi(\mathbf{x}))(D^{-1}A\phi(\mathbf{x}))^T
    \label{eq:relationalKernMean}
    =D^{-1}AK_XAD^{-1}$.$
$

In contrast to the propositional kernel mean, the convergence of the relational to its population counterpart is not necessarily guaranteed because of sample dependence. 
We discuss convergence and consistency guarantees under the assumption of weak dependence after describing the relational independence tests.

\subsection{Relational Marginal Independence Test}
With the relational kernel mean defined we now turn to the central task of this paper, non-parametric inference of relational dependence~(NIRD). 
As a test statistic, we use the Hilbert-Schmidt independence criterion~(HSIC)~\citep{gretton-icalt05}.
HSIC measures the maximum distance between an embedding of the observed joint distribution, and the product of the marginals, i.e., 
$
    \|\mathbb{E}[\phi(x)\otimes\phi(y)]  - \mathbb{E}[\phi(x)]\otimes\mathbb{E}[\phi(y)]\|^2
$. 
We perform a hypothesis test using HSIC as the test statistic where the null hypothesis refers to independence. The test produces a p-value which is used to decide whether to reject the null or not. Testing relational independence using HSIC is straightforward with the relational kernel mean by using the kernel matrix defined earlier
in the empirical HSIC estimator. 
Defining the centering matrix $H = I - \frac{1}{n}\mathbf{1}\mathbf{1}^\top$, an empirical estimate of HSIC is given by $\frac{1}{n^2}\text{trace}\left({K_XHK_YH}\right)$, where $K_X$ and $K_Y$ are kernel matrices corresponding to the random variables $X$ and $Y$, respectively. 
Independence testing with HSIC 
can be performed by using the corresponding relational kernel in the test statistic.

\subsection{Relational Conditional Independence Test}
A similar construction can be employed to test for relational conditional independence, defined in Definition \ref{dfn:reldep_cond}.
Following \citet{strobl-jci19}, we consider the following $L^2$ spaces, 
\begin{align*}
&F_{X Z} \triangleq\left\{\tilde{f} \in L_{X Z}^{2} \mid E(\tilde{f} \mid Z)=0\right\} \\
&F_{Y Z} \triangleq\left\{\tilde{g} \in L_{Y Z}^{2} \mid E(\tilde{g} \mid Z)=0\right\} \\
&F_{Y \cdot Z} \triangleq\left\{\tilde{h}^{\prime} \mid \tilde{h}^{\prime}=h^{\prime}(Y)-E\left(h^{\prime} \mid Z\right), h^{\prime} \in L_{Y}^{2}\right\}
\end{align*}

Each of these quantities can easily be constructed by considering regressions, e.g. $\tilde{f}$ can be obtained by taking the residuals after performing a regression. 
We consider a mean of the feature basis representation as an aggregation function whenever one of the variables is relational. 
Under the assumption that the direct sum of the reproducing kernel Hilbert spaces, $k_x k_y$ and $k_z$ is dense in $L_2$, \citet{strobl-jci19}~(proposition 5) showed that conditional linear covariance of zero implies uncorrelatedness, i.e., 
$\mathbb{E}\left[\tilde{f}\tilde{g}\right] = 0 \implies X \indep Y | Z \implies \Sigma_{XY | Z} = 0$.
This motivates the use of a multiple output kernel ridge regression as an estimator of the conditional expectation, $\beta = (\phi(z)^T\phi(z) + \lambda I)^{-1}\phi(z)^T\phi(\ddot{x})$ {where $\ddot{X} \triangleq (X,Z)$ is the concatenation of x and z}. {Informally, this can be seen as applying the ``{kernel trick}'' of considering linear operations on non-linear transformations of the data allowing for observations to be dependent}. 
The test is then constructed by considering the residuals, $\widetilde{\phi(\ddot{x})} = \phi(\ddot{x}) - \phi(z)\beta_{\ddot{x}z}, \widetilde{\phi(y)} = \phi(y) - \phi(z)\beta_{yz}$ and sum of the squared covariances between them. The final form of the test is given by $\frac{1}{n^2}\text{trace}\left({\widetilde{K_{\ddot{X}}}H\widetilde{K_Y}H}\right)$ where 
$\widetilde{K_{\ddot{X}}}$ and $\widetilde{K_Y}$ refers to the kernel matrices for the residuals $\widetilde{\phi(\ddot{x})}$ and $\widetilde{\phi(y)}$ respectively.

There are two considerations in employing this procedure in a relational setting, namely how to handle relational variables in the conditioning set ($Z$) and the test set ($X$), respectively.
When a member of the conditioning set is relational, the test procedure is identical after replacing $\phi(z)$ with its relational counterpart, $\frac{1}{|\hat{\mathcal{N}}(z)|}\sum_{m \in \hat{\mathcal{N}}(z)} \phi(m)$.
When a member of the test set is relational, the problem is reduced to predicting each member of the set independently by considering the regression of the perspective of the relational variable, as described by \citet{maier-uai13}.
After regressing individual members, the mean of residuals is then considered for the marginal tests, $\widetilde{\sigma(\phi(x))} = \frac{1}{|\hat{\mathcal{N}}(x)|}\sum_{m \in \hat{\mathcal{N}}(x)} \phi(m) - \phi(z)\beta_{m z}$.

\subsection{Consistency of Relational Independence Test}
\label{subsec:consistency}

In order to reason about the behavior of test statistics under non-i.i.d. samples and understand asymptotic behavior we need to characterize the behavior of dependence amongst instances as a function of some notion of distance between instances. 
There are a number of formalisms for reasoning about dependent data~\citep{andrews-jstor94,bickel-jstor99,dedecker-book07}. 
In this work we focus on weak dependence~\citep{dedecker-book07}, which we describe next.

\subsubsection{Weak Dependence}
In order to accommodate dependent observations and maintain consistency of the testing procedure we will assume that observations are weakly dependent.
Weak dependence provides a flexible notion of dependence that requires only the definition of distance between instances and the presence of a measurable probability space. Within this work we will make use of the notion of weak dependence, i.e. $\tau$-dependence.

\begin{dfn}\citep{dedecker-book07}
Let $\boldsymbol{\pi}$ be a filtration\footnote{A filtration is an ordering of a set such that for any two subsets, $S_{1,\dots,j}, S_{1,\dots,k}$, $j \leq k$ $\rightarrow S_{1,\dots,j} \subseteq S_{1,\dots,k}$.} over the set of nodes in a graph, $G$, defined by performing a breadth first search at an arbitrary node, $v \in G$. 
Further, define $X$ to be a $\mathbb{L}^p$-integrable random variable. 
The \textbf{weak-dependence coefficient} is defined as $\tau_{p,r}(X) = \sup_{(i,j)}\|\sup_g \text{Cov}(g\left(X_{\boldsymbol{\pi}(i)}\right), g\left(X_{\boldsymbol{\pi}(j)}\right))\|_p$, where $i \leq j$ and $j - i \leq r$, and $g()$ is a Lipschitz function.

\end{dfn}
Intuitively, the weak dependence coefficient, $\tau_{p,r}(X)$ measures the covariance between a vector, $X_i$ and another random vector $X_j$ drawn from the same process separated by at least distance of $r$.
We call a process \textit{weakly dependent} if $\tau$ tends to zero as the $r$ tends to infinity. 
Note that this is a strictly weaker condition than alternative assumptions on dependence such as strong mixing and $m$-dependence which require independence at a finite distance, whereas weak-dependence only requires it asymptotically.

\subsubsection{Weak Dependence in Relational Domains}

We provide a natural extension of weak dependence within the relational setting by replacing usual definition of distance to the shortest path distance between two nodes in a graph. 
The role of $\tau$ in this case can be interpreted as measuring the decay of dependence between instances as a function of shortest-path distance. 
We will assume from here  out that as the distance between any two nodes in the network tends to infinity, the dependence between them converges to zero.
More formally, we will employ the following assumption:
\begin{assumption}
$(X_t)_{t\in\pi}$ is a strictly stationary $\tau$-dependent process with $\sum_{r=1}^\infty r^2\sqrt{\tau_r(X)} \leq \infty$ for some filtration $\pi$, where $r$ is shortest-path graph distance. 
\end{assumption}

The notion of weak dependence within the network setting is not novel to this work,
\citet{xiang-aistats11} make use of the $\tau$-coefficient in the context of deriving asymptotic consistency for transductive learning
with an assumption of linear dependence amongst instances. 
However, to our knowledge, our work is the first to consider weak dependence with arbitrary dependence for independence testing of relational data. Consistency of the relational independence testing is provided by the following theorem and corollary, after applying two additional assumptions.

\begin{assumption}
\label{assump:finite}
The maximum degree of any node in the network is bounded by a real constant. 
\end{assumption}

\begin{assumption}
\label{assump:fixed}
The network structure is fixed and doesn't change during the generation of the observed random variables.
\end{assumption}
Assumption \ref{assump:finite} ensures that the average shortest path distance from any node to all other nodes in the graph tends to infinite as the number of nodes tends to infinite, which is necessary in order to have convergence of weakly dependent sequences. 
Assumption \ref{assump:fixed} ensures that the observed neighborhoods for nodes correspond to the structure which generated the data. 

\begin{theorem}
\label{thm:consprop}
Under the aforementioned assumptions the Hilbert-Schmidt independence criterion of two weakly dependent propositional variables converges in $L_1$ to its population counterpart, i.e., $\left|\text{HSIC}_n -\text{HSIC}_{\text{population}}\right| \underset{d}{\longrightarrow}0$.
\end{theorem}

\begin{corrollary}
\label{cor:consprop}
Under the aforementioned assumptions the Hilbert-Schmidt independence criterion between a weakly relational and a weakly dependent propositional variable converges in $L_1$ to its population counterpart, i.e., $\left| \overline{\text{HSIC}_n} -\text{HSIC}_{\text{population}}\right| \underset{d}{\longrightarrow}0$.
\end{corrollary}

The proof of theorem \ref{thm:consprop}, which is deferred to the supplement, follows by observing that the empirical estimate of HSIC is a degenerate $V$-statistic and then through a proof which shows consistency of degenerate $V$-statistics under weak-dependence in structured domains, which may be of independent interest. Similarly, the proof of the corollary, also deferred to the supplement follows from theorem \ref{thm:consprop} and showing that the weak dependence coefficient remains finite for relational variables. 

It is important to note that Theorem \ref{thm:consprop} and Corollary \ref{cor:consprop} show convergence in distribution but do not claim any guarantees regarding the rates of convergence with respect to the number of nodes and level of dependence. The rate of convergence will depend on the weak dependence coefficient. In the case that the coefficient is 0, this reduces to results that correspond to prior work on iid data~\citep{zhang-uai11}. While there is prior work studying this in more restrictive assumptions on the dependence between instances~\citep{london-pmlr13}, we are not aware of similar results for the case of weak dependence in general structured domains even in the simpler case of regression. This would be an important direction for future work.

\subsection{Large Scale Approximations}
While the proposed model is theoretically appealing, the associated time and space complexity render it infeasible for most modern network settings. 
To address this, we appeal to an approximation of the kernels known as Random Fourier Features~\citep{rahimi2008random}. 
Random Fourier Features exploit Bochner's theorem, which states that a continuous, time-invariant kernel is positive definite if and only if the kernel is the Fourier transform of some non-negative measure. 
For example the Gaussian kernel can be represented with the following Fourier transformation $\hat{k}(\omega)=\frac{1}{2 \pi} \int e^{-j \omega^{\top} \delta} k(\delta) d \delta$.
This property implies that a kernel can be approximated via the following procedure: 
\begin{itemize}[leftmargin=*]
    \item Draw $D$ samples, from some distribution (i.e Normal), to approximate the Gaussian kernel where the variance $\sigma$ corresponds to the bandwidth of the kernel.
    \item Construct the Fourier basis explicitly as {
        $z\left(x\right)=
        \sqrt{\frac{2}{d}}\left[\cos \left(w_{1}^{T} x\right), \sin \left(w_{1}^{T} x\right), \ldots, \right]$.
    }
    \item Perform linear operations using $z$.
\end{itemize}
Following \citep{zhang2018large,strobl-jci19}, we approximate HSIC using random Fourier features by considering 
$
    \widehat{\text{HSIC}}(X, Y) = \left\|\frac{1}{n} \boldsymbol{Z}_{X}^{T} \boldsymbol{H} \boldsymbol{Z}_{Y}\right\|^{2}
$
where $Z$ is a $n \times d$ dimensional matrix with each row consisting of the random Fourier features for an observation. 
We can represent the relational kernel mean as $D^{-1}AZ$, and the corresponding test statistic as $\|\frac{1}{n}Z_X^TAD^{-1}HZ_Y\|$ where $D$ and $A$ are the diagonal degree and adjacency matrix as before. In several experiments we show  that using approximate statistic leads to significant performance improvements with minimal effect on the efficacy of the test, even with only a few random features.

\section{Experiments}
\label{sec:exp}

We run experiments with multiple network datasets, relational dependence cases, and synthetic attribute generators to evaluate the effectiveness of the proposed test. %

\subsection{Network datasets}
\label{subsec:data}
We consider networks from two synthetic graph generators and three non-PII real-world networks. %
First, for the \text{\ba} (BA) model, we vary the parameter that controls the number of nodes a new node can attach to. For the \text{\er} (ER) model, we vary the probability of edge creation between each pair of nodes. For each set of parameters, we generate $100$ networks with size 100. The small size of the synthetic networks is driven by the baseline method which does not scale well, as shown in Figure \ref{sfig:scale}. %
We also demonstrate the applicability of our approach through a Facebook ego-network with $4,039$ nodes and $88,234$ edges~\citep{leskovec-nips12}. %
The other two real-world datasets (\emph{Twitter}, \emph{50 Women}) and corresponding experimental results are described in the Appendix. %

\subsection{Four cases of relational (in)dependence}
We choose three representative relational dependence cases and one relational independence case to cover a range of possible tests. %
We consider attributes $Z, X, Y\in \bm{\mathcal{A}}$  which measure characteristics in time steps $t-1, t, t+1$ respectively. %
All the cases are represented with arrows showing the direction of dependence: 

\begin{enumerate}[leftmargin=*]
    \item \textbf{Case 1: $\rel{X} \rightarrow v_i.Y$}
    \item \textbf{Case 2: $\rel{X} \leftarrow v_i.Z \rightarrow v_i.Y \leftarrow \rel{X}$}
    \item \textbf{Case 3: $v_i.X \leftarrow \rel{Z} \rightarrow v_i.Y \leftarrow v_i.X$}
    \item \textbf{Case 4: $\rel{X} \leftarrow v_i.Z \rightarrow v_i.Y$}
\end{enumerate}

where $\rel{X}$ and $\rel{Z}$ are relational variables on the attributes $X$ and $Z$ of the direct neighbors of $v_i$. %
Case 1 refers to marginal independence between a relational and a propositional variable ($\rel{X} \indep v_i.Y$). 
Cases 2 and 3 introduce conditional independence given a confounder. Case 2 refers to a propositional confounder ($\rel{X} \indep v_i.Y | v_i.Z$) whereas case 3 refers to a relational confounder ($v_i.X \indep v_i.Y | \rel{Z}$). %
A test should be able to reject the null hypothesis of no dependence in the first three cases. Case 4 represents conditional independence and the test should not reject the null hypothesis and it should produce high errors. Note that direction is ignored in the test. The synthetic attribute generation process is described in Appendix.

\begin{figure*}[h!]
    \centering
    \subfloat{\includegraphics[width=.76\textwidth]{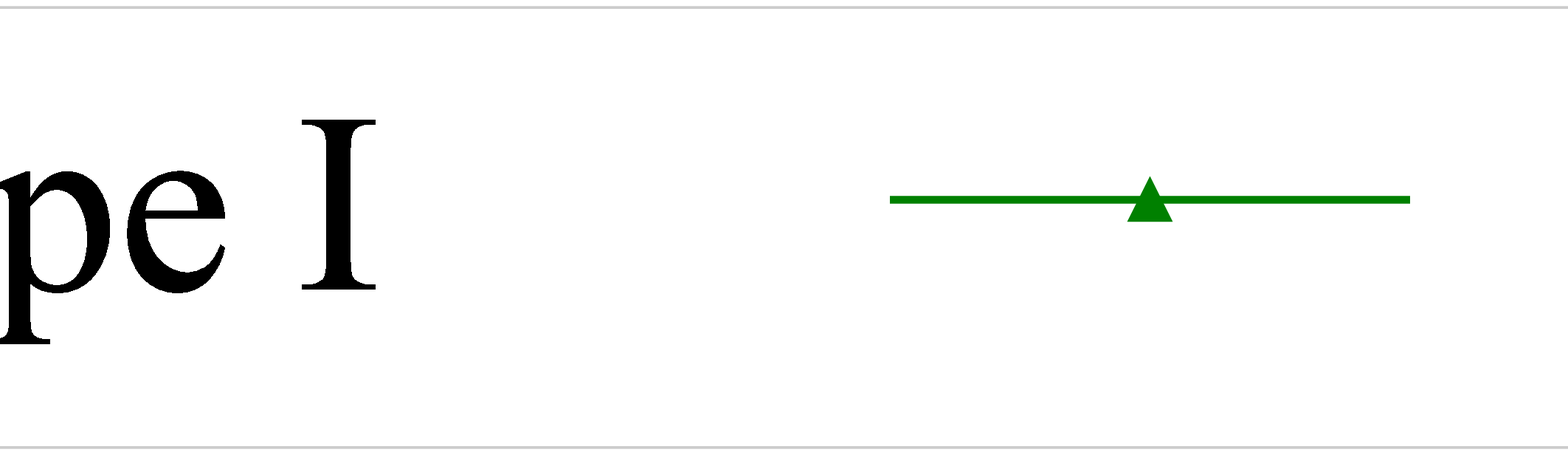}}\\
    
    \setcounter{subfigure}{0}
    
    \subfloat[BA: Case 1 ]{\label{sfig:dep_s_a}\includegraphics[width=0.28\textwidth]{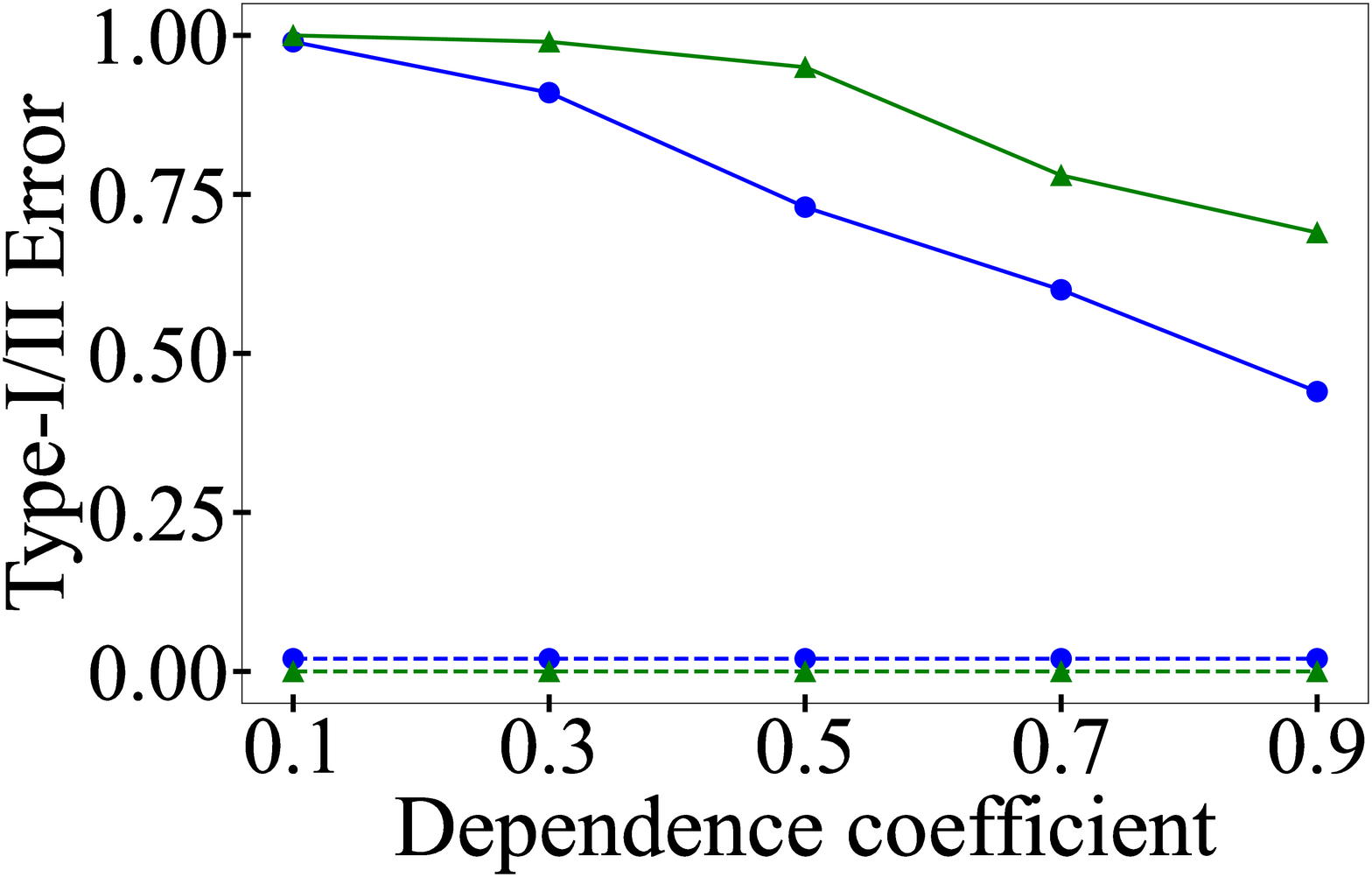}}
    \hspace{1em}
    \subfloat[BA: Case 2 ]{\label{sfig:dep_s_c}\includegraphics[width=0.28\textwidth]{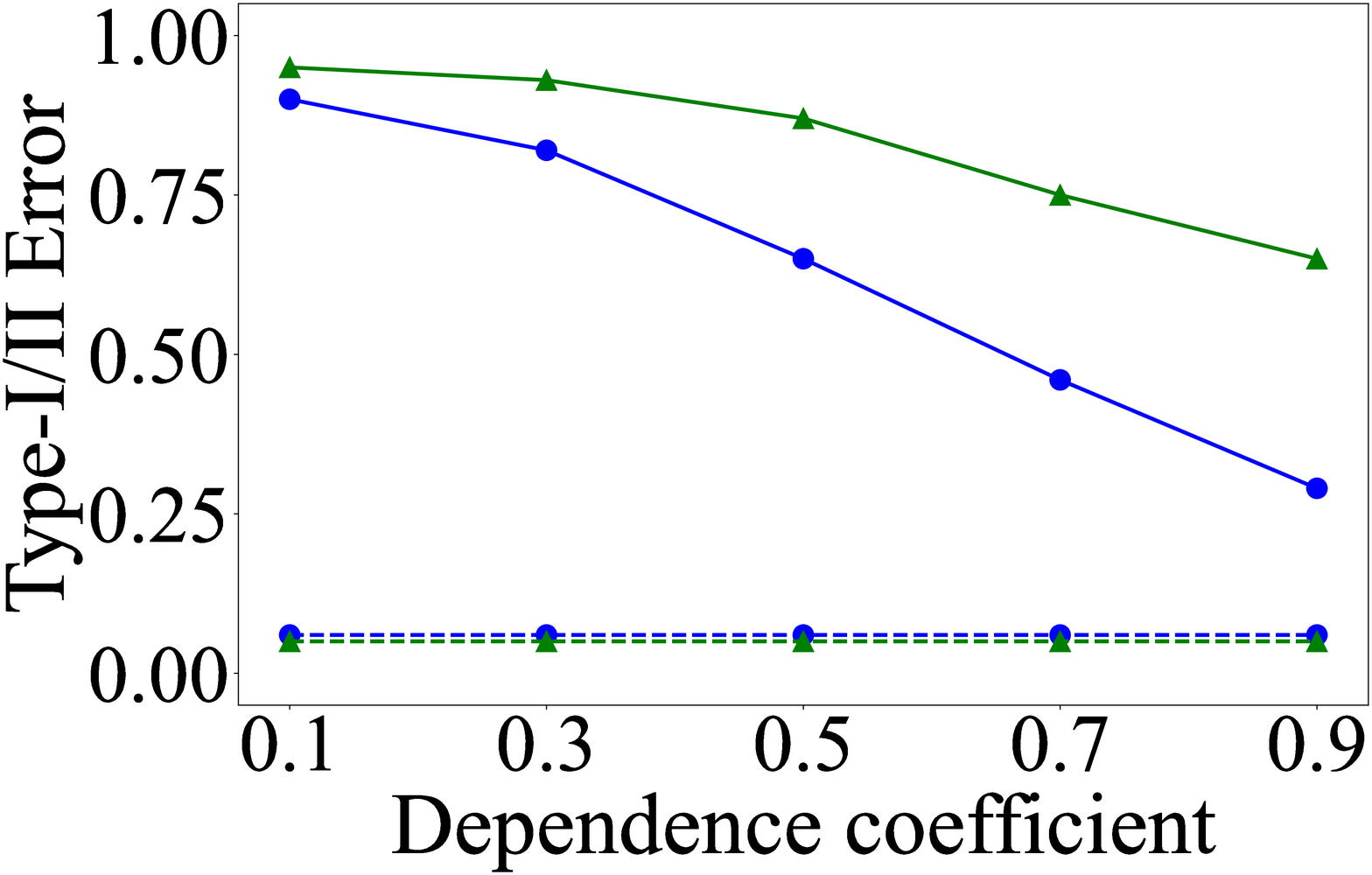}}
    \hspace{1em}
    \subfloat[BA: Case 3 ]{\label{sfig:dep_s_e}\includegraphics[width=0.28\textwidth]{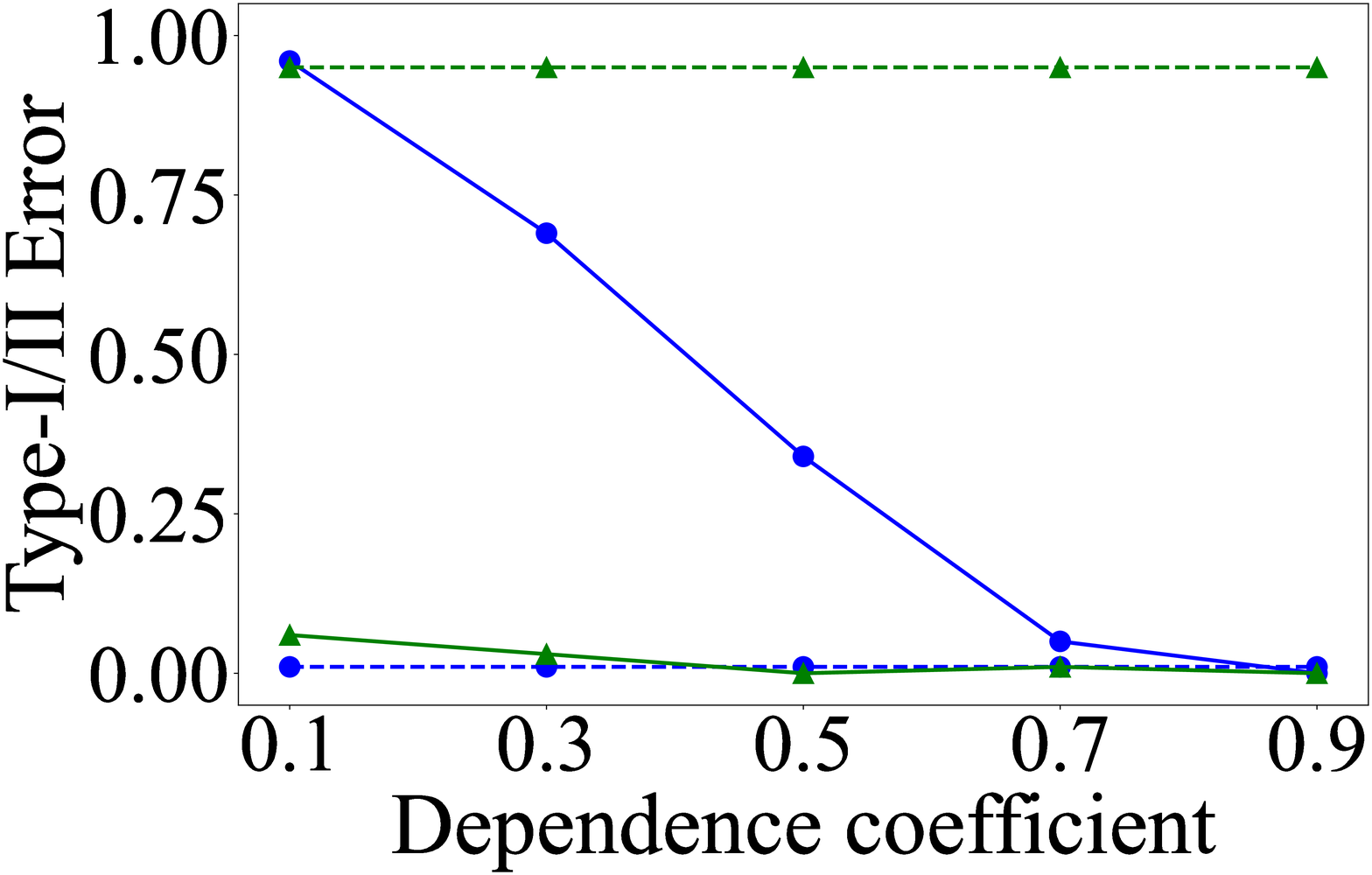}}\\
    
    \subfloat[ER: Case 1 ]{\label{sfig:dep_s_b}\includegraphics[width=0.28\textwidth]{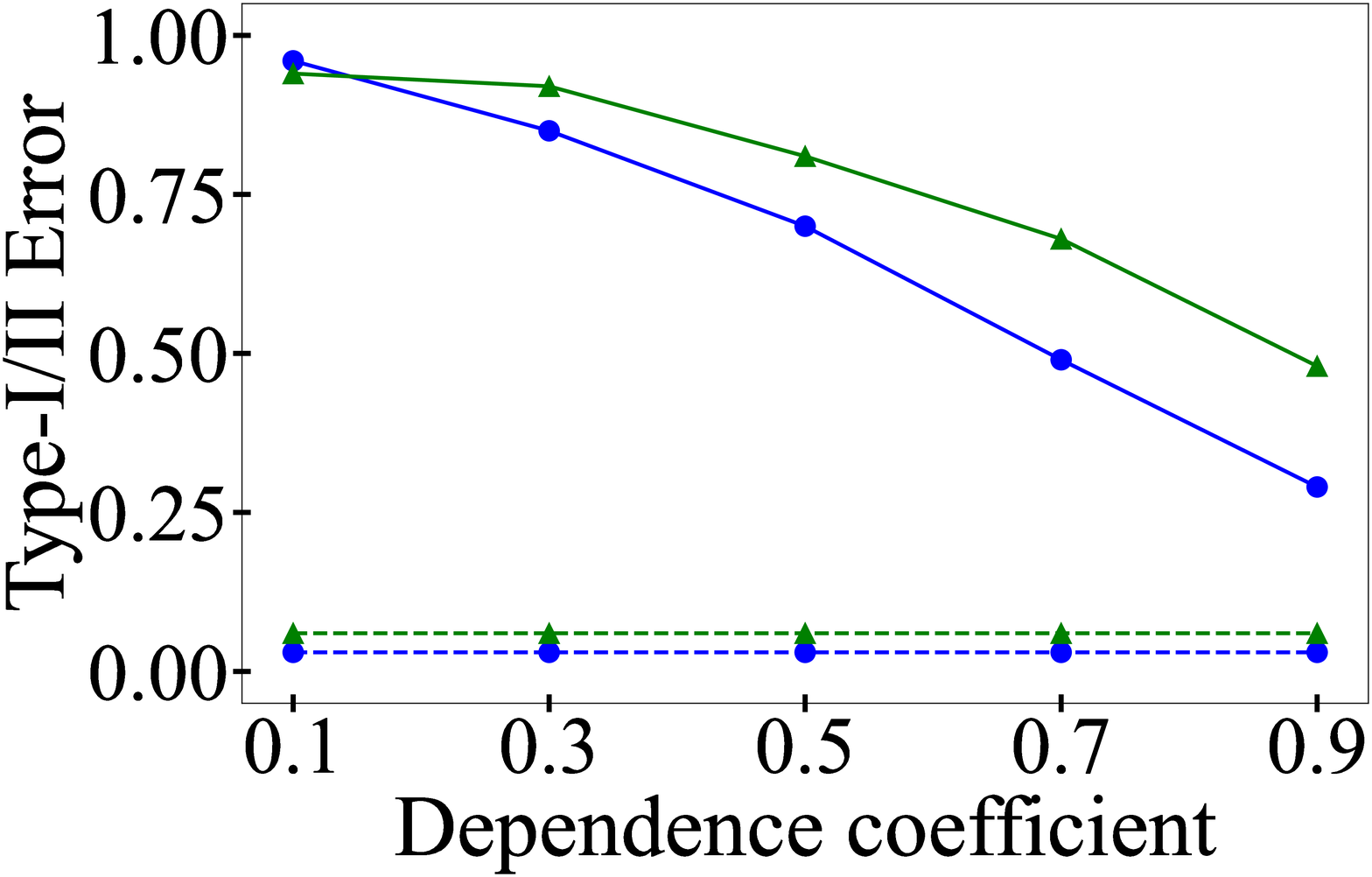}}
    \hspace{1em}
    \subfloat[ER: Case 2 ]{\label{sfig:dep_s_d}\includegraphics[width=0.28\textwidth]{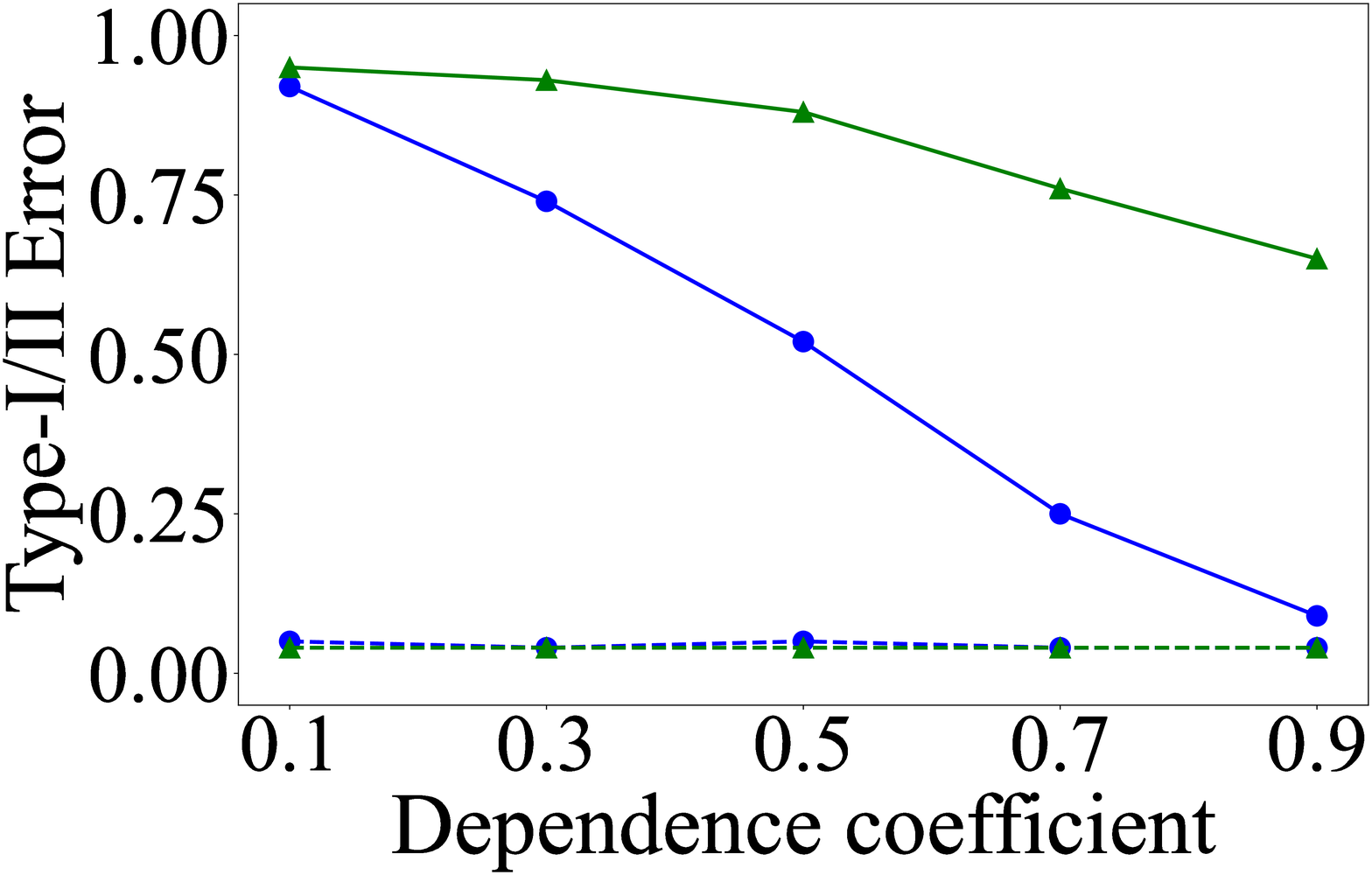}}
    \hspace{1em}
    \subfloat[ER: Case 3 ]{\label{sfig:dep_s_f}\includegraphics[width=0.28\textwidth]{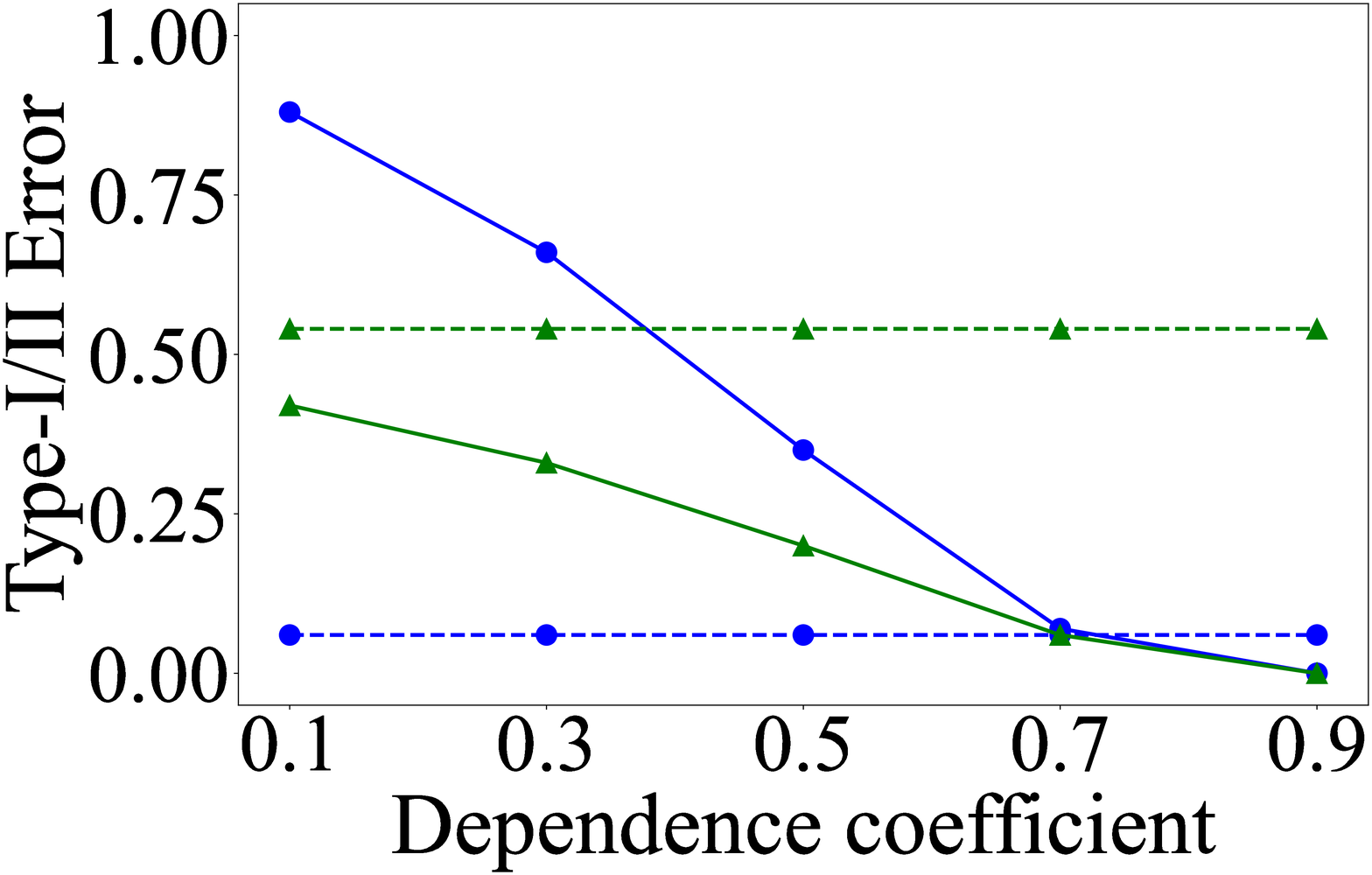}}\\

    \caption{Relational dependence impact on Type I/II errors.}%
    
    \label{fig:dep_s}
\end{figure*}

\subsection{Experimental Setup}
We empirically evaluate the proposed approach, \text{\mcit}, to the state-of-the-art RCI test method, KRCIT~\citep{lee-uai17}. \footnote{Code available at https://github.com/edgeslab/nird-uai22} We report the average Type I and Type II errors with significance level 0.05 over 100 trials for each set of parameters. We use \textit{Radial Basis Function kernel} (RBF) as the base kernel. 
KRCIT is implemented with HSIC as the kernel-based marginal independence test method and KCIT~\citep{zhang-uai11} as the kernel-based conditional independence (CI) test. We use the approximate method of NIRD in all experimental evaluation with $20$ and $50$ random Fourier features for marginal and conditional test respectively. 
We estimate the null distribution via permutation on the non-relational variable since the marginal distribution remains unchanged.
We compare both RCI methods (NIRD, KRCIT) to a recent i.i.d. CI test method, Sobolov Independence Criterion (SIC)~\citep{mroueh-nips19} (see Appendix). 
We study NIRD's strengths and weaknesses in five experimental setups:

\textbf{Relational dependence sensitivity}: We evaluate the sensitivity of the dependence tests to different relational dependence strengths. We report results on polynomial models while varying the dependence coefficient in range $\{0.1, 0.3, 0.5, 0.7, 0.9\}$ for the alternate hypothesis and report both Type I, II errors. Edge connectivity is $3$ for BA and edge probability is $0.02$ for ER model.

\textbf{Diffusion}: We apply NIRD to test for contagion by simulating a linear threshold model~\citep{granovetter-ajs78} with initial treatment probability of $0.1$ on the semi-synthetic Facebook network. We reassign treatment values in each diffusion step and generate outcomes ($Y$) based on treatments ($X$) generated in the last diffusion step. The attribute generation process is described in the Appendix. We expect the distribution of $\rel{X}$ to change with increasing diffusion steps and investigate at what step it is possible to detect relational dependence. We vary the number of steps, sample size and measure Type II error. %
We also investigate the impact of activation probability on the Type-II error on the Twitter ego-network with 10,000 nodes (results in the Appendix).

\textbf{Network sensitivity}: We examine performance over a variety of network structures. We vary edge connectivity of BA in range $\{1, 2, 3\}$ and edge probability of ER in range $\{0.005, 0.01, 0.015, 0.02, 0.025\}$. We use a fixed dependence coefficient value of $0.5$.

\textbf{Scalability}: To compare the scalability of NIRD against the baseline KRCIT, we generate \text{\er} synthetic networks (edge probability 0.02) with varying number of nodes and report their execution time for marginal and conditional independence testing. We vary the network size (x-axis) in the range $\{100, 200, 300, 400, 500\}$. The choice of small networks was driven by the fact that KRCIT scales exponentially as shown in Figure \ref{sfig:scale} and it is impossible to run for larger networks. In the Appendix, we also demonstrate the scalability of NIRD through the diffusion experiment which is run on networks of size up to 10k nodes. The experiment is run on a 2.4GHz 8-core machine with 50GB memory.  %

\textbf{Real world demonstration}: In the Appendix, we demonstrate the applicability of our test for detecting peer influence in a well-studied real-world social network (\emph{50 Women}) where our test discovers smoking-, drug- and sport-related peer dependencies that concur with previous research.

\subsection{Results}

\begin{figure}[h!]
	\centering
	\subfloat[Linear threshold model]{\label{sfig:ltm}\includegraphics[width=.32\textwidth]{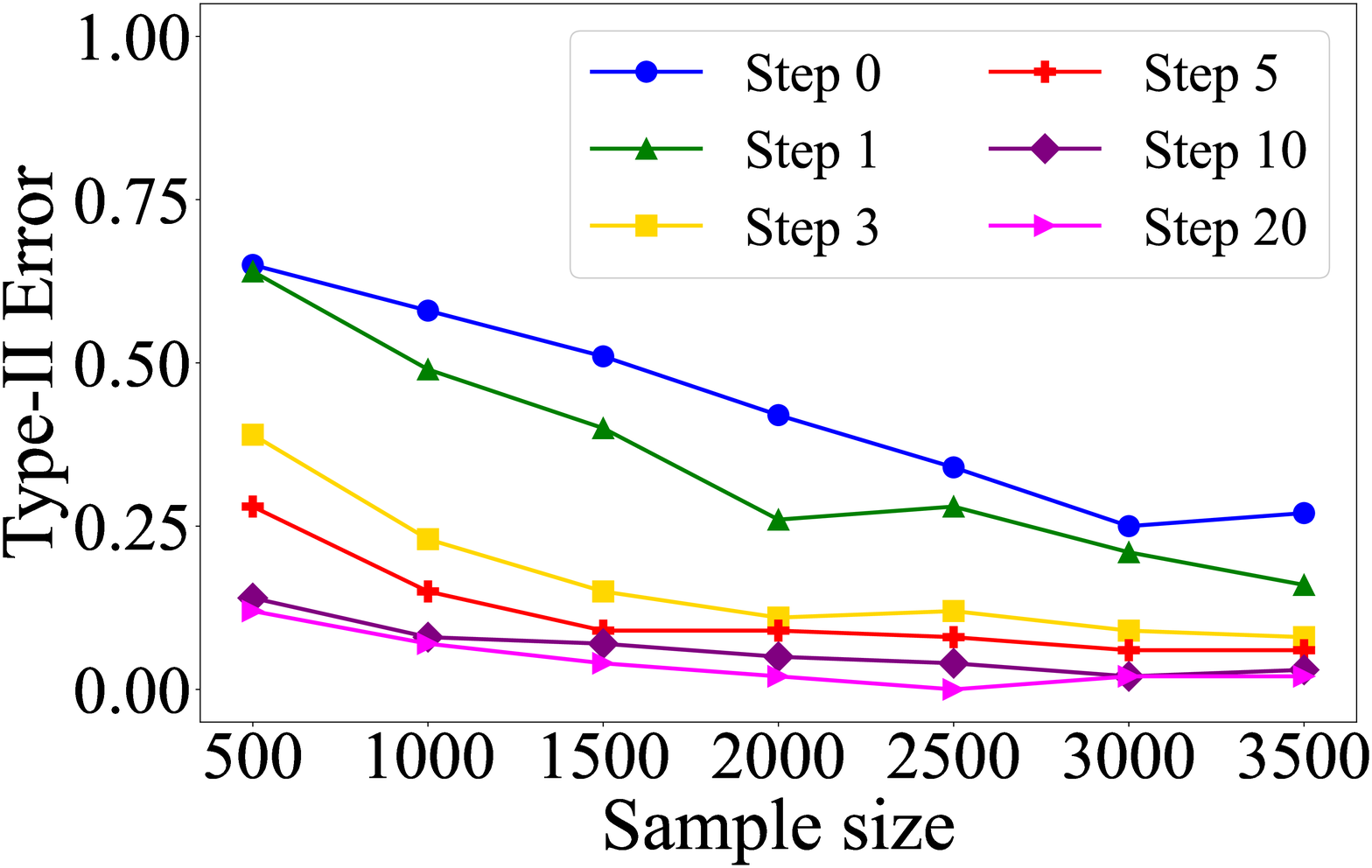}}
	
	\caption{Impact of sample size on Type II error.}
\end{figure}

\textbf{Relational dependence sensitivity}. Figure \ref{fig:dep_s} shows Type I and Type II errors for the polynomial dependency model on synthetic data. The rows correspond to the network models and the columns to relational dependence cases. The solid and dashed lines correspond to Type II and Type I errors respectively. The test is most challenging when the dependence coefficient, $\beta_d$ (x-axis) is low. The figure shows that both RCI methods are well calibrated with low Type I error (max 0.06 by KRCIT in \text{\er}) for the first two cases. In these cases, NIRD consistently produces lower Type II errors compared to KRCIT. It is most visible in \text{\er} model (\ref{sfig:dep_s_d}) with $86\%$ reduction in Type II error for $\beta_d = 0.9$. The performance gain of NIRD increases slightly from case 1 to case 2 as the difficulty increases. In case 3, KRCIT is poorly calibrated and exhibits an unusually high Type I error. Across cases, NIRD shows desired behavior: it is consistently well-calibrated and its Type II error decreases with the increase of relational dependence. Case 4 provides a sanity check and both methods produce high Type II errors (0.9 to 1.0) with good calibration. The error is nearly constant irrespective of strength of dependence coefficients or network model parameters used. In order to test for sensitivity to noise, we repeat these experiments varying the noise variance over multiple trials instead of drawing from a fixed distribution. The results look very similar (see Appendix).

\begin{figure*}[h!]
	\centering
	\subfloat{\includegraphics[width=.76\textwidth]{fig/fig_1_2_legend.eps}}\\
	
	\setcounter{subfigure}{0}
	
	\subfloat[BA: Case 1 ]{\label{sfig:net_a}\includegraphics[width=0.28\textwidth]{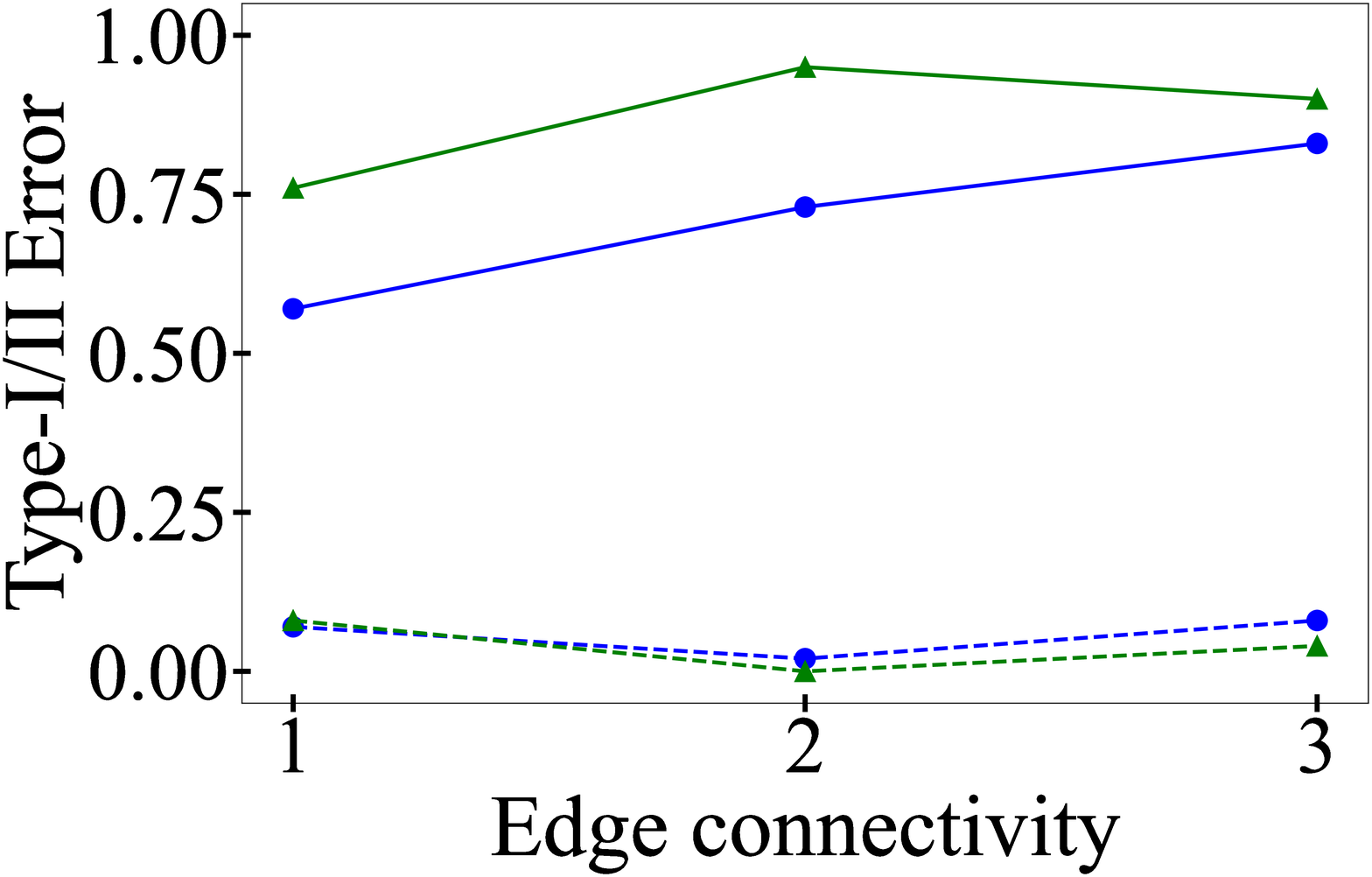}}
	\hspace{1em}
	\subfloat[BA: Case 2 ]{\label{sfig:net_c}\includegraphics[width=0.28\textwidth]{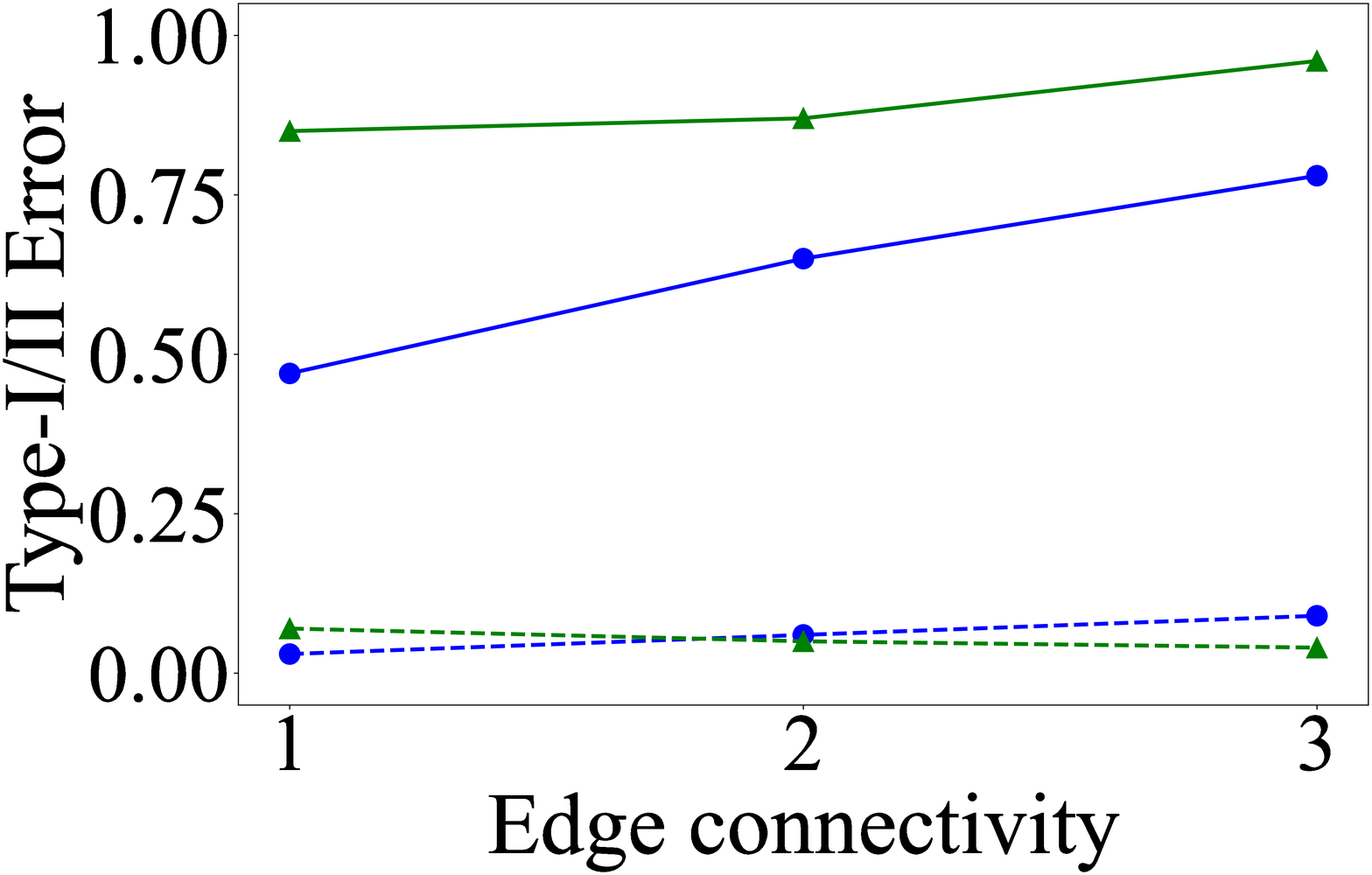}}
	\hspace{1em}
	\subfloat[BA: Case 3 ]{\label{sfig:net_e}\includegraphics[width=0.28\textwidth]{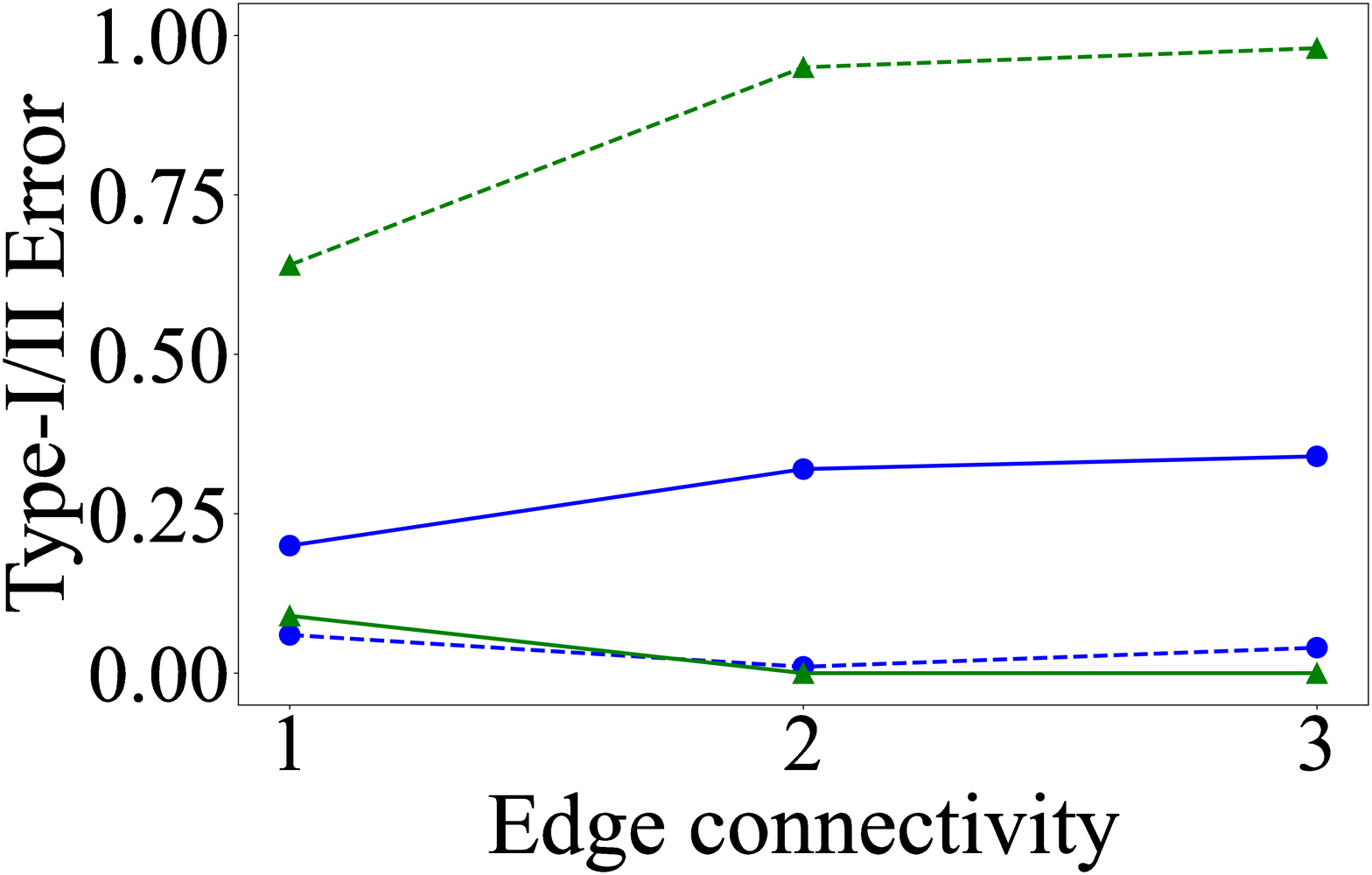}}\\
	
	\subfloat[ER: Case 1 ]{\label{sfig:net_b}\includegraphics[width=0.28\textwidth]{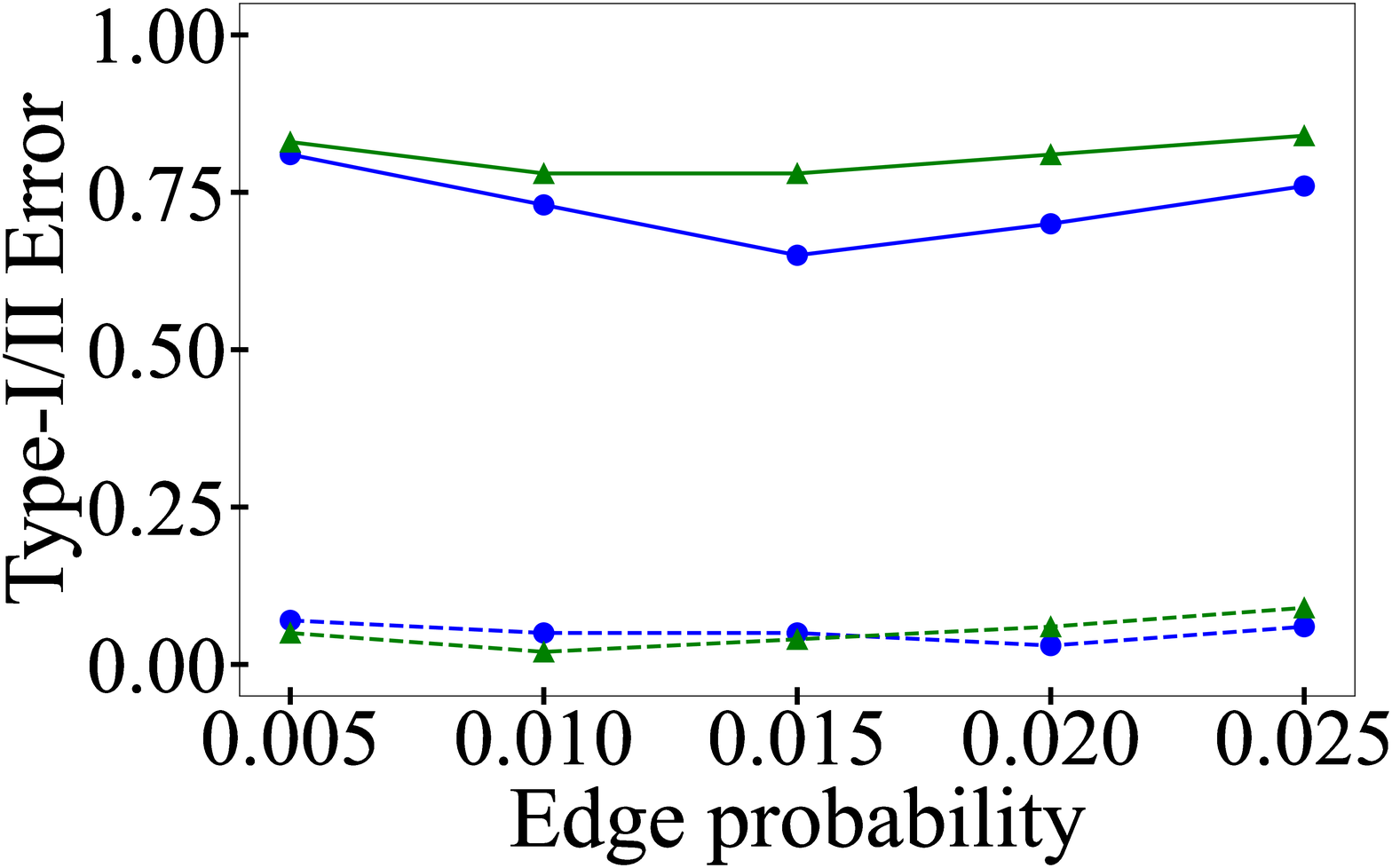}}
	\hspace{1em}
	\subfloat[ER: Case 2 ]{\label{sfig:net_d}\includegraphics[width=0.28\textwidth]{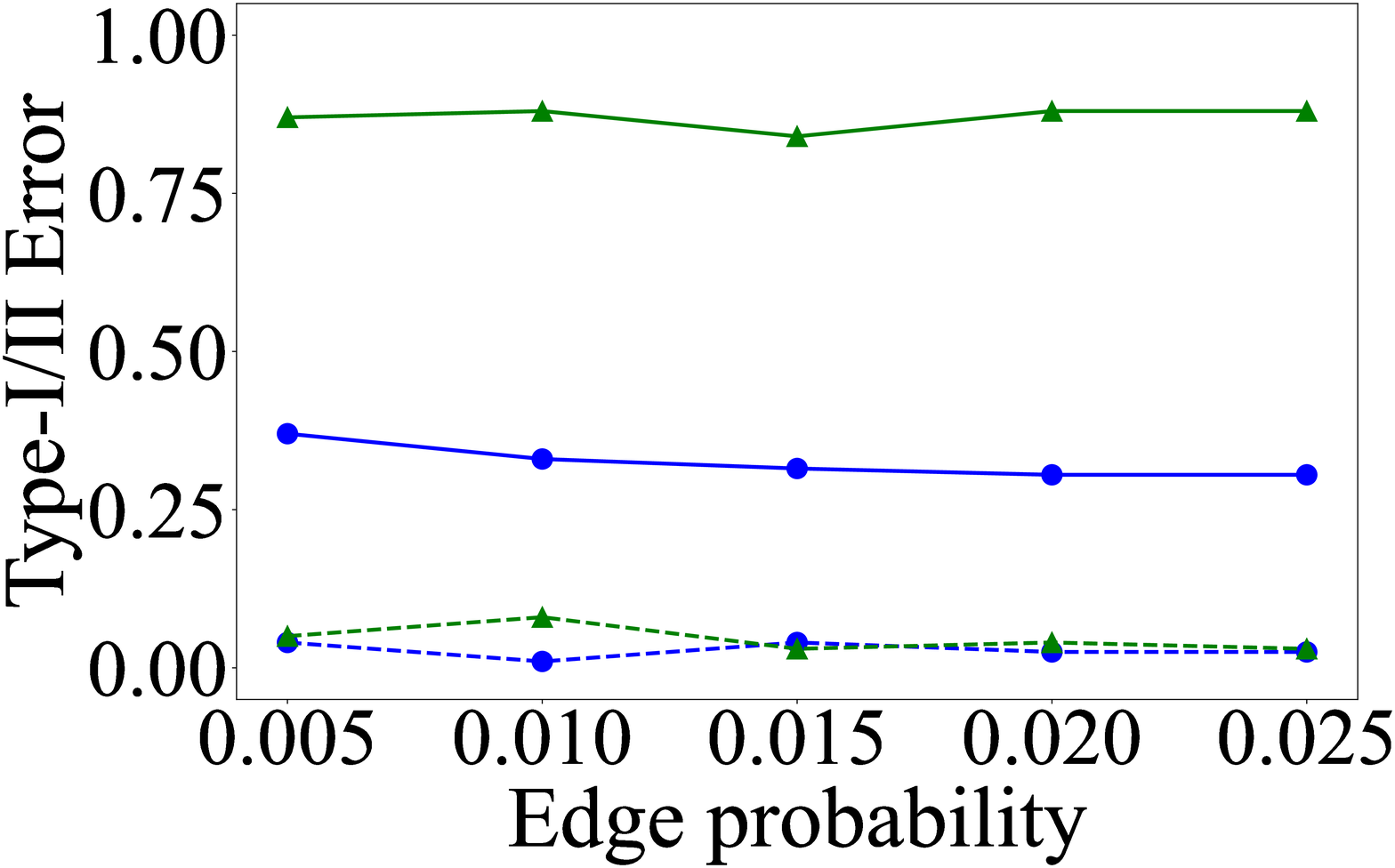}}
	\hspace{1em}
	\subfloat[ER: Case 3 ]{\label{sfig:net_f}\includegraphics[width=0.28\textwidth]{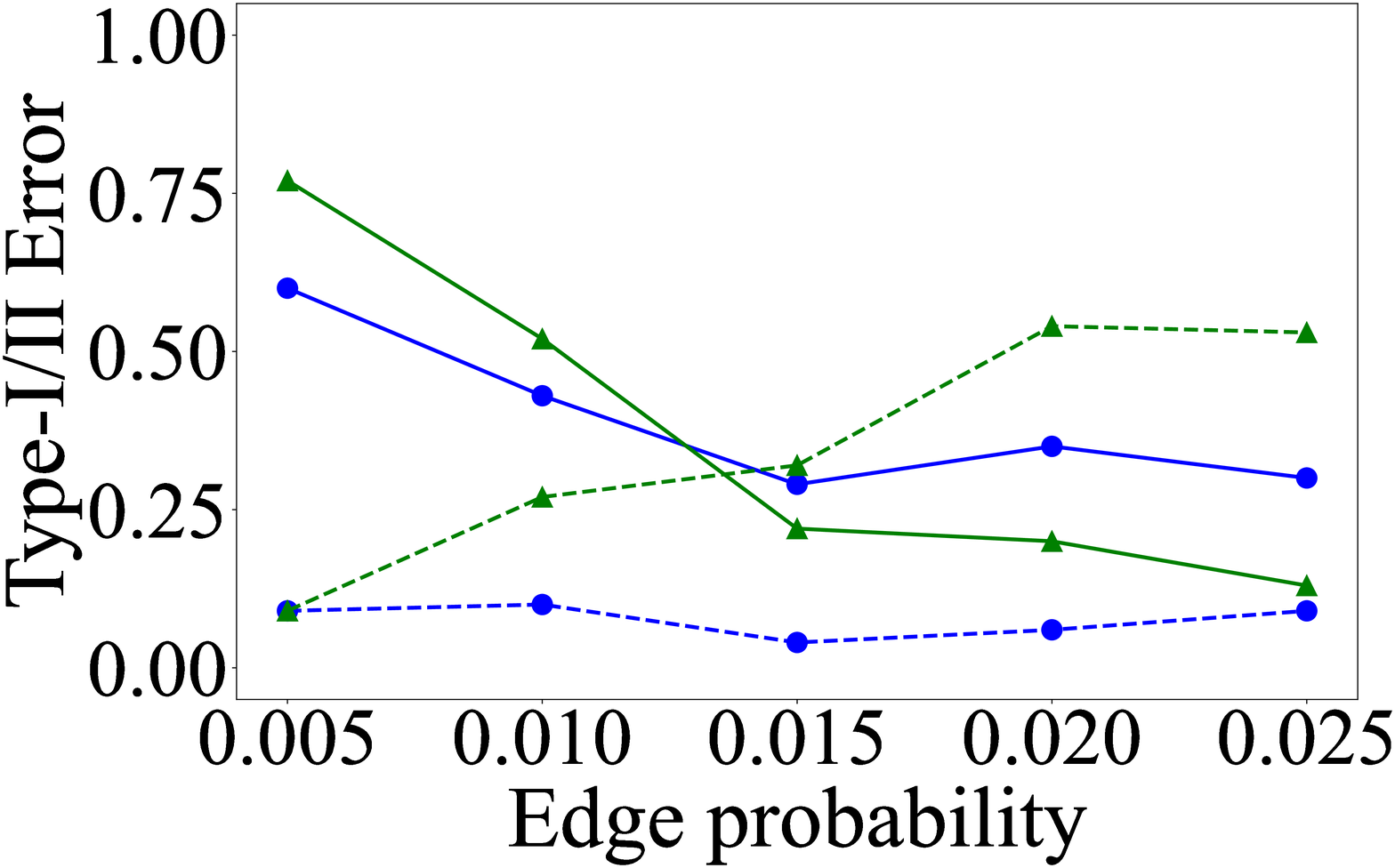}}\\

	\caption{Impact of network parameters on Type I/II errors.} %
	
	\label{fig:net_models}
\end{figure*}

\textbf{Diffusion}.
Figure \ref{sfig:ltm} shows the impact of the number of diffusion steps (lines) and sample size (x-axis) on the effectiveness of NIRD. 
At initial activation (1 diffusion step) there is a high Type II error across sample sizes which decreases with higher number of steps. %
We see a significant decrease in error with just 5 diffusion steps. Further steps drastically lower the Type II error and at 20 steps and larger samples it can reject the null hypothesis consistently. This suggests that relational dependence is easier to detect after several diffusion steps rather than at early activation. It also demonstrates the effectiveness and scalability of NIRD in terms of detecting social phenomena in real world networks. Note that it is computationally infeasible to run the baseline method on such size of samples.

\textbf{Network sensitivity}. %
Figure \ref{fig:net_models} shows Type I and Type II errors for two network models. The x-axis represents the corresponding parameter values for each model. We observe that increased parameter values exhibit higher Type II errors in general for \text{\ba} model but not for \text{\er}. A possible reason is that \text{\ba} exhibits a more skewed degree distribution compared to \text{\er}. Note that the increased parameter values indicate higher density of the network. We expect \text{\er} to show a similar trend if the edge probability is further increased.
\text{\mcit} outperforms KRCIT in terms of Type II error (except in Figures \ref{sfig:net_e} and \ref{sfig:net_f} which is due to poor calibration of the baseline method) irrespective of network density. Type II error is reduced as high as $65\%$ for \text{\er} model with edge probability $0.025$ (Figure \ref{sfig:net_d}). Moreover, Type I error for NIRD is consistent whereas KRCIT suffers in case 2 (Figures \ref{sfig:net_e}, \ref{sfig:net_f}).

\begin{figure}[h!]
    \centering
    \subfloat[Test scalability]{\label{sfig:scale}\includegraphics[width=.32\textwidth]{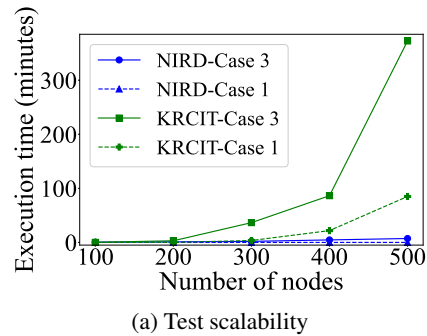}}\hfill
    
    \caption{Impact of sample size on execution time.}
\end{figure}

\textbf{Scalability}.
Figure \ref{sfig:scale} shows execution time in minutes (y-axis) for both marginal (case 1) and conditional (case 3) independence test for different network sizes in terms of number of nodes (x-axis). %
The solid and dashed lines represent the  conditional and marginal test result respectively. KRCIT exhibits an exponential complexity whereas NIRD shows much less sensitivity to network size. This is expected given the complexity of the corresponding algorithms. %

\section{Conclusion}
In this work we examine the problem of defining and measuring statistical dependence in relational data. 
We propose NIRD, a consistent, non-parametric test for detecting relational dependence that improves state-of-the-art relational dependence testing by capturing a wide range of possible relational dependencies. Moreover, we introduce an approximate method that 
makes NIRD scalable to larger networks.
We evaluate the effectiveness of our method across diverse relational settings and find that our proposed test
exhibits significantly less sensitivity to network properties and dependence types. 
Our work paves the way for a number of promising future research directions, from testing for social influence to causal structure learning from relational data. %

\section{Acknowledgments}

This material is based on research sponsored in part by NSF under grant No.\@ 2047899, and DARPA under contract number HR001121C0168.

\bibliography{ahsan_640}

\newpage
\clearpage
\appendix
\onecolumn

\section{Proofs}
In this section we present the proofs of consistency for HSIC and relational HSIC under weak dependence. 
The approach here is to extend the results of \citet{chwialkowski-nips14} and \cite{leucht-jma13}, who analyze degenerate $U$ and $V$-statistics (which includes HSIC as a specific instantiation) under weak dependence in spaces that admit euclidean distances to the more general setting of graph structured spaces.
Much of the results carry through after modifications to accommodate the fact that the number of reachable instances at a specific distance is irregular. We present a modification of the relevant proof which shows the convergence of the distribution of degenerate $V$-statistics, which may be of independent interest, and then describe the application to our setting and the extension to relational variables. 

\subsection{$V$-Statistics Under Relational Weak Dependence} 
Let $X = \{X_1,\dots X_n\}$ be the set of given observations. 
Define $h$ to be a symmetric function, taking $m$ arguments. 
A $V$-statistic is a function defined with respect to $h$ taking the form

\begin{align*}
    V(h, X)_n = \frac{1}{n^m}\sum_{i\in i_1 \dots i_m \in N^m} h(X_{i_1}, \dots X_{i_m})
\end{align*}

where $N^m$ is defined as the Cartesian product of the set $1,\ldots,n$ and $n$ is the total number of observations. 
In the sequel, we will write $V(h,X)$ as $V(X)$ to reduce notational clutter. 
We will refer to $h$ as the \textit{core}\footnote{In order to prevent confusion, we follow \citet{chwialkowski-nips14} and do not follow the canonical convention of calling $h$ the kernel.}. 

We say that a core $h$ is \emph{$j$-degenerate} if for every $x_1,\dots,x_j$, 
\begin{align*}
E[h(X_1,\dots,X_j,X^*_{j+1},\dots,X^*_m)] = 0
\end{align*}
where
$X^*_{j+1},\dots,X^*_m$ are independent samples drawn from the same distribution as $X_1$.
A core is called canonical if for all $j \leq m - 1$ it is $j$-degenerate. 
Finally, we call a $V$-statistic with a 1-degenerate core a \emph{degenerate $V$-statistic}.

We now provide a proof of consistency of degenerate $V$-statistics for relational data under weak dependence. 
The strategy of this proof is to first approximate the $V$-statistic with weighted sums of squares, and then apply the central limit theorem to this approximation. 
The approximation used is the spectral decomposition of the core
\begin{align*}
 h(x, y) = \sum_k\lambda_k\Phi(x)\Phi(y)
\end{align*}
where $\lambda_k$ are the nonzero eigenvalues of $E[h(x, X_0)\Phi(X_0)] = \lambda\Phi(x)$, and $\Phi(x)$ are the associated eigenvectors.
This strategy largely mirrors what is found by \citet{leucht-jma13}.
However, in that case, the approximations are constructed as a function of distance in time. 
Our contribution is a generalization of the approximations to network domains that follow the aforementioned assumptions.
This is done by considering \textit{sets} of instances separated by shortest path distance of $k$, rather than assuming that there is always a single instance at distance $k$, and adapting results accordingly.

\setcounter{theorem}{1}
\begin{theorem}
\label{thm:weakv}
Let $(Z_k)_k$ be centered, jointly normal random variables with $\Cov(Z_j, Z_k) = \sum_{r=-\infty}^\infty\Cov(\Phi_j(X_0), \Phi_k(X_r))$, and $(\lambda_k)_k, (\Phi_k)_k$ be the sequence of non-zero eigenvalues and corresponding eigenfunctions of 
$E\left[h(x, X_0)\Phi(X_0)\right] = \lambda\Phi(x)$.
Under the aforementioned assumptions, $V_n \overset{d}{\longrightarrow} Z := \sum_k \lambda_k Z^2_k$, as $n \rightarrow \infty$, and 
$EZ = \sum_{r\in\mathbb{Z}}Eh(X_0,X_r) < \infty$
i.e., the infinite series that defines $Z$ converges in $L_1$. 
\end{theorem}

\begin{proof}
Let $(\lambda_k)_k$ be an enumeration of the positive eigenvalues of $Eh(x,X_0)\Phi(X_0)=\lambda\Phi(x)$ sorted in decreasing order, and $(\Phi_k)_k$ be the corresponding eigenfunctions. 
Following ~\citet{leucht-jma13}, we set $\lambda_k := 0, \Phi_k \equiv 0, \forall k > L$, when the number $L$ of non-zeros eigenvalues is finite. 
We are given from a version of Mercer's theorem (given by Theorem 2 of Sun~\cite{sun-jc05}) that 
\begin{equation*}
    \label{eq:series}
    h^{(K)}(x,y) = \sum_{k=1}^K\lambda_k\Phi_k(x)\Phi_k(y) \underset{K\rightarrow\infty}{\longrightarrow}h(x,y), \forall x,h \in \textrm{supp}(P^{X_0}) 
\end{equation*}
\citet{leucht-jma13} provide the prerequisites necessary for the equation to converge absolutely and uniform on compact subsets of $\textrm{supp}(P^{X_0})$, which apply directly in our setting as well. 
We will consider an approximation of $V_n$ by a $V$-statistic with a kernel with finite spectral decomposition given by $V_n^{(K)} = \frac{1}{n}\sum_{s,t}^n h^{(K)}(X_s, X_t)$. 
Because $h$ is positive semi-definite by definition, all eigenvalues are non-negative, implying $V_n - V_n^{(K)} \geq 0$. 
This implies
\begin{align*}
    &E\left| V_n - V_n^{(K)}\right| = E\left[ V_n - V_n^{(K)}\right]\\
    &=E\left[ h(X_0, X_0) - h^{(K)}(X_0, X_0) \right] + \sum_{r=1}^{n-1}2(1-r/n)E\left[ h(X_0, X_r) - h^{(K)}(X_0, X_r) \right]
\end{align*}
By majorized convergence the first term converges to zero as $K\rightarrow\infty$.
For the second term, repeated application of Cauchy-Schwarz gives
\begin{align*}
    &\sum_{r=1}^{n-1}2(1-r/n)E\left[ h(X_0, X_r) - h^{(K)}(X_0, X_r) \right]\\
    &\leq 2\sum_{r=1}^\infty\left| \sum_{j \in \Delta_r} E\left[ \sum_{k=K+1}^\infty \lambda_k \Phi_k(X_0)\Phi_k(X_j) \right] \right|\\
    &= 2 \sum_{r=1}^\infty  \left| E\left[\sum_{j \in \Delta_r} \sum_{k=K+1}^\infty \lambda_k \Phi_k(X_0)   (\Phi_k(X_j) - \Phi_k(\widetilde{X}_j)) \right] \right|\\
    &\leq 2\sum_{r=1}^\infty \sqrt{E\left[\sum_{j \in \Delta_r}  \sum_{k=K+1}^\infty \lambda_k\Phi_k^2(X_0) \right]} \sqrt{E\left[  \sum_{j \in \Delta_r}\sum_{k=K+1}^\infty \lambda_k\left( \Phi_k(X_r) - \Phi_k(\widetilde{X}_j) \right)^2 \right]}\\
    &\leq 2\sqrt{\sum_{r=1}^\infty \lambda_k}\sum_{r=1}^\infty\sqrt{E\left[ \sum_{j \in \Delta_r}\sum_{k=1}^\infty \lambda_k \left( \Phi_k(X_j) - \Phi_k(\widetilde{X}_j) \right)^2 \right]}\\
    &\leq 2\sqrt{\sum_{k=K+1}^\infty\lambda_k}\sum_{r=1}^\infty\sqrt{\sum_{j \in \Delta_r} E \left[ h(X_j, X_j) - h(X_j, \widetilde{X}_j) - h(\widetilde{X}_j, X_j) + h(\widetilde{X}_j, \widetilde{X}_j) \right]}\\
    &\leq 2\sqrt{\sum_{k=K+1}^\infty}\sum_{r=1}^\infty\sqrt{2\max(\textrm{deg})^r\textrm{Lip(h)}}\sqrt{\tau(r)}
\end{align*}
Where $\Delta_r$ is the set of nodes whose shortest path distance from $X_0$ is $r$, $\max(\textrm{deg})$ is the largest degree in the network, and $\widetilde{X}_r$ denotes a copy of $X_r$ that is independent of $X_0$ and satisfies $E\|X_r - \widetilde{X}_r\|_1 \leq \tau(r)$. 
Because $\sum_{k=1}^\infty\lambda_K = Eh(X_0, X_0) < \infty)$, thus $\sum_{k=K+1}^\infty\lambda_k \rightarrow 0$ as $K\rightarrow\infty$ we arrive at $\underset{n}{\sup}E\left| V_n - V_n^{(K)} \right| \underset{K\rightarrow\infty}{\longrightarrow} 0 $. 

The proof of the central limit theorem for partial sums, i.e., for $K \leq L$
\begin{align}
    V_n^{(K)} = \sum_{k=1}^K\lambda_k\left(n^{-1/2}\sum_{t=1}^n \Phi_k(X_t) \right)^2 \overset{d}{\longrightarrow} \sum_{k=1}^K \lambda_kZ^2_k
\end{align}
follows a direct application of ~\cite{leucht-jma13} Theorem 2.1 proof part (\textit{ii}). 
Combining these two results, to satisfy the requirements of Theorem 2 of ~\citet{dehling-spa09} we arrive at $V_n \overset{d}{\longrightarrow} Z := \sum_k\lambda_kZ^2_k$. 
The only item remaining to be shown is $EX < \infty$, which follows from a direct application of part (\textit{iv}) of the proof of Theorem 2 provided by ~\citet{leucht-jma13}. 
\end{proof}

We now turn our attention to the Hilbert-Schmidt independence criterion.
Note that both follow almost immediately from implications of theorem \ref{thm:weakv}.

\setcounter{theorem}{0}

\begin{theorem}
\label{thm:consprop}
Under the aforementioned assumptions the Hilbert-Schmidt independence criterion of two weakly dependent propositional variables converges in $L_1$ to its population counterpart, i.e., $\left|\overline{\text{HSIC}_n} -\text{HSIC}_{\text{population}}\right| \underset{d}{\longrightarrow}0$.
\end{theorem}
\begin{proof}
Recall that the Hilbert-Schmidt independence criterion~(HSIC) is a test of dependence, i.e. a hypothesis test of paired samples where the null hypothesis is that the two samples are generated independently, $\mathbb{P}_{x,y} = \mathbb{P}_x\mathbb{P}_y$. 
Our focus is on the empirical estimator of HSIC, which can be written as degree-four $V$-statistic with a core defined by:
\begin{align}
\label{eq:vsic}
    h(x_1, x_2, x_3, x_4) =
    &\frac{1}{4!}\sum_{\pi\in S_4} k(x_{\pi(1)}, x_{\pi(2)})k(y_{\pi(1)}, y_{\pi(2)}) + 
    k(y_{\pi(3)}, y_{\pi(4)})
    - 2k(y_{\pi(2)}, y_{\pi(3)})
\end{align}
where $S_n$ is the set of permutations over a set of $n$ elements.
Convergence then follows as a direct application of theorem \ref{thm:weakv} and the weak law of large numbers.
Note that under independence $Z$ is a zero mean, jointly Gaussian variable and the resulting sequence $\sum_i \lambda_i Z^2_i$ is mean zero.
\end{proof}

\setcounter{corrollary}{0}
\begin{corrollary}
Under the aforementioned assumptions the Hilbert-Schmidt independence criterion between a weakly relational and a weakly dependent propositional variable converges in $L_1$ to its population counterpart, i.e., $\left|\text{HSIC}_n -\text{HSIC}_{\text{population}}\right| \underset{d}{\longrightarrow}0$.
\end{corrollary}
\begin{proof}
The central items to be shown in order to apply the results of theorem \ref{thm:weakv} to apply are (1) relational kernels define a valid $V$-statistic, and (2) the relational variable remains weakly-dependent. 
Item (1) follows directly by denoting one of the variables in equation \ref{eq:vsic} to be a set of instances return by the path predicate and $k$ to be the relational kernel defined in the main text. 
Item (2) follows as a consequence of assumption 5 which bounds the degree of each node by a finite constant, $c$. As a result, any path predicate which defines a finite length path will return a set no larger than $c < c' < \infty$. As a result, so long as the initial random variable is weakly dependent, the relational variable constructed from the initial random variable will also be weakly-dependent, albeit with a slower rate of convergence since the coefficient $\tau_r$ (the weak dependence coefficient) will necessarily decay more slowly. 
\end{proof}

\section{Extension to Multi-relational Systems}

In our problem definition we assumed a single-entity, single relationship relational schema for ease of exposition. Here, we discuss necessary extensions for a multiple entity, multi-relational system. We consider a set of item classes $\bm{\mathcal{I}}$ to be the union of entities and relationship classes, $\bm{\mathcal{I}} = \bm{\mathcal{E}} \cup \bm{\mathcal{R}}$, following prior work ~\citep{lee-uai17,maier-uai13}. We refer to the attribute class of an item class $I \in \bm{\mathcal{I}}$ as $\bm{\mathcal{A}}(I)$. Moreover, let $G(I)$ denote a set of items of an item class $I \in \bm{\mathcal{I}}$.

Here, we point out two major differences in a multi-relational system:

\begin{enumerate}
    \item The relational dependence is specifically defined between two item classes $I \in \bm{\mathcal{I}}$ and $I \in \bm{\mathcal{J}}$.
    \item The path predicate $\rho$ is likely to be defined with relational queries rather than random walks over a %
    neighborhood.
\end{enumerate}

Now, we revisit definition 1 from the main text with the new notation as follows:

\begin{dfn}[Relational Variable]
Given a relational schema $\mathcal{S} = \langle \bm{\mathcal{E}}, \bm{\mathcal{R}}, \bm{\mathcal{A}} \rangle$, its instantiation $G$, two item classes $I,J \in \bm{\mathcal{I}}$ and a path predicate $\rho$, a relational variable $\sigma(v_i, \bm{X}, G, \rho)$ is the set of attributes $v_j.\bm{X}$ selected by $\rho$ of items $v_j \in G(J)$ reachable from items $v_i \in G(I)$ such that $\bm{X} \subset \bm{\mathcal{A}}(J)$, where the path predicate $\rho$ is a function given by:

\[
    \rho(v_i, G) : G(I) \mapsto \mathcal{P}(G(J))
\]

\end{dfn}

The necessary assumptions and relational dependence definitions still hold. The major difference arises in the compact representation of the relational kernel. Equation 1 stays valid with an updated notion of path predicate. However, the compact representation in equation 2 is no longer trivial since the adjacency matrix $A$ is no longer directly applicable. There are two potential workarounds. First, since the compact representation is not mandatory for our method to work, we can still work with equation 1 for multi-relational systems. Second, we can essentially consider the bipartite graph between sets of items between item classes $I, J \in \bm{\mathcal{I}}$ and use the adjacency matrix $A_{IJ}$ of this bipartite graph instead of $A$. Similarly a corresponding degree matrix $D_{IJ}$ can be constructed from $A_{IJ}$.

\section{Experiments}
\subsection{Synthetic Attribute Generation} \label{subsec:att_gen}
Here, we describe the synthetic attribute generation procedure for the three cases mentioned in the main text. Note that, only the generation of $v_i.Y$ differs in null and alternate hypothesis while others stay the same. We consider polynomial dependency model for most of our experiments. $v_i.X$ for case 1 and $v_i.Z$ for cases 2,3 is drawn from a uniform distribution $U(0, 1)$ while $v_i.X$ is always \textit{binarized} to resemble the effect of treatment assignment. The outcome $v_i.Y$ is generated according to the following equation for marginal dependence (case 1):
\begin{equation}
    v_i.Y \thicksim
            \begin{cases}
              U(0, 1) & null\\
              \beta_d \cdot (g(\rel{x}))^2  + \epsilon & alternate
            \end{cases}
\end{equation}

Conditional dependence (case 2) is reflected by the following equation:
\begin{equation}
\label{eqn:case1}
    \begin{split}
        & v_i.X \thicksim \beta_c \cdot (v_i.Z)^2  + \epsilon\\
        & v_i.Y \thicksim
            \begin{cases}
              \beta_c \cdot (v_i.Z)^2  + \epsilon & null\\
              \beta_d \cdot (g(\rel{X}))^2 + \beta_c \cdot (v_i.Z)^2 + \epsilon & alternate
            \end{cases}    
    \end{split}
\end{equation}

Here, $\beta_d$ and $\beta_c$ are dependence and confounding coefficients respectively. $\beta_c$ is considered 1.0 in our experiments. $\epsilon$ is noise drawn from standard normal ($N(0, 1)$) distribution. $g$ refers to the \textit{mean} aggregate function. We can get the generating function for case 3 by replacing $g(\rel{X})$ and $v_i.Z$ with $v_i.X$ and $g(\rel{Z})$ respectively in equation \ref{eqn:case1}. Next, we consider the following procedure to simulate linear threshold model for the diffusion experiment which falls under case 1:
\begin{equation}
    \begin{split}
        T_i & \thicksim U(0, 1)\\
        v_i.x_{t+1} & = \mathbbm{1}(mean(\rel{x_t}) > T_i)\\
        v_i.y_{t+1} & = \mathbbm{1}(g(\rel{x_t}) > T_i)
    \end{split}
\end{equation}
where we reassign $v_i.x$ values to simulate each diffusion step based on its value in previous step. The $v_i.y$ values are assigned based on $v_i.x$ values in the last diffusion step.

\setcounter{figure}{4}

\begin{figure*}[h]
    \centering
    \subfloat{\includegraphics[width=.76\textwidth]{fig/fig_1_2_legend.eps}}\\
    
    \setcounter{subfigure}{0}
    
    \subfloat[BA: Case 1 ]{\label{sfig:vnv_a}\includegraphics[width=0.28\textwidth]{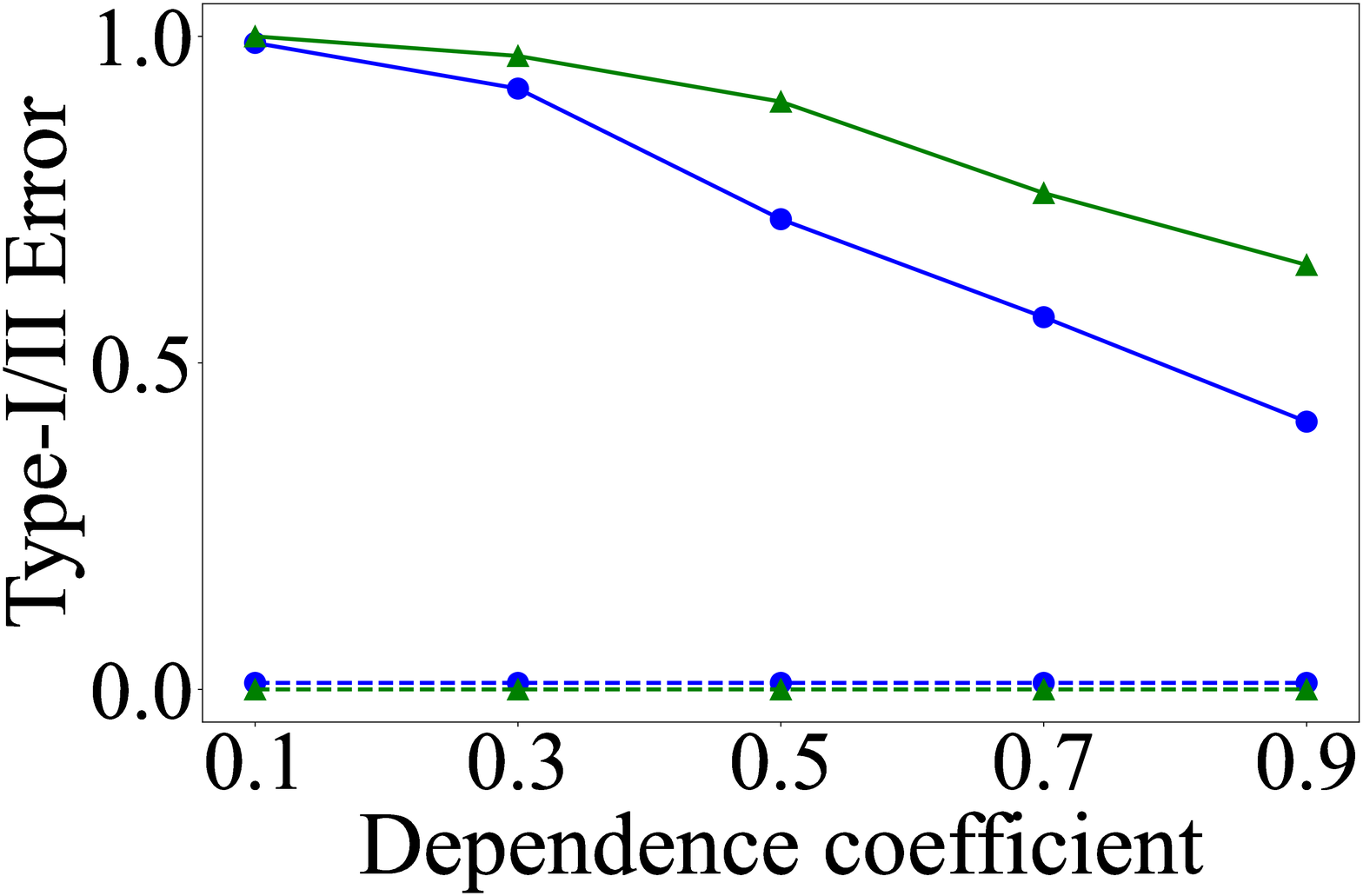}}
    \hspace{1em}
    \subfloat[BA: Case 2 ]{\label{sfig:vnv_c}\includegraphics[width=0.28\textwidth]{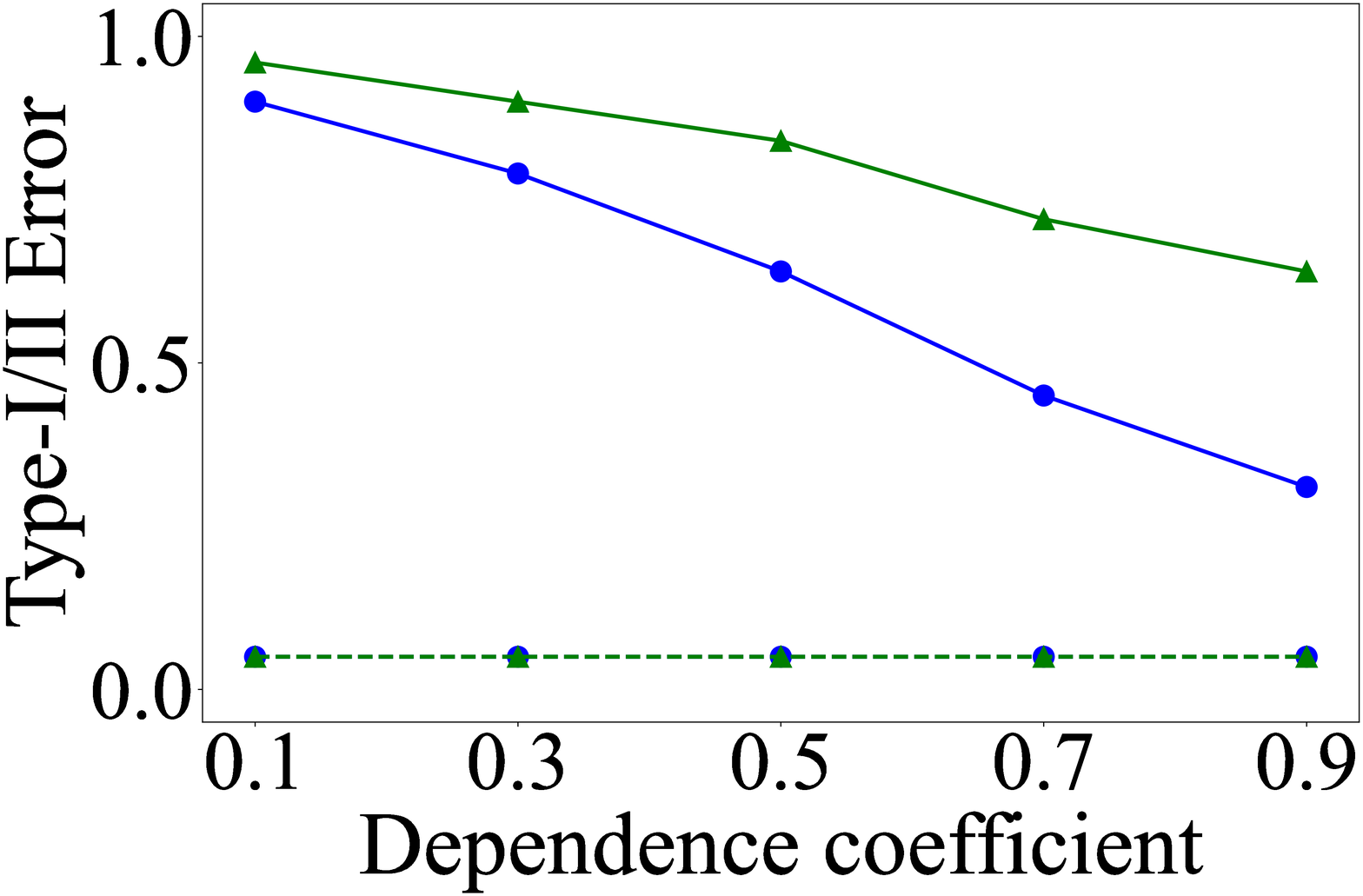}}
    \hspace{1em}
    \subfloat[BA: Case 3 ]{\label{sfig:vnv_e}\includegraphics[width=0.28\textwidth]{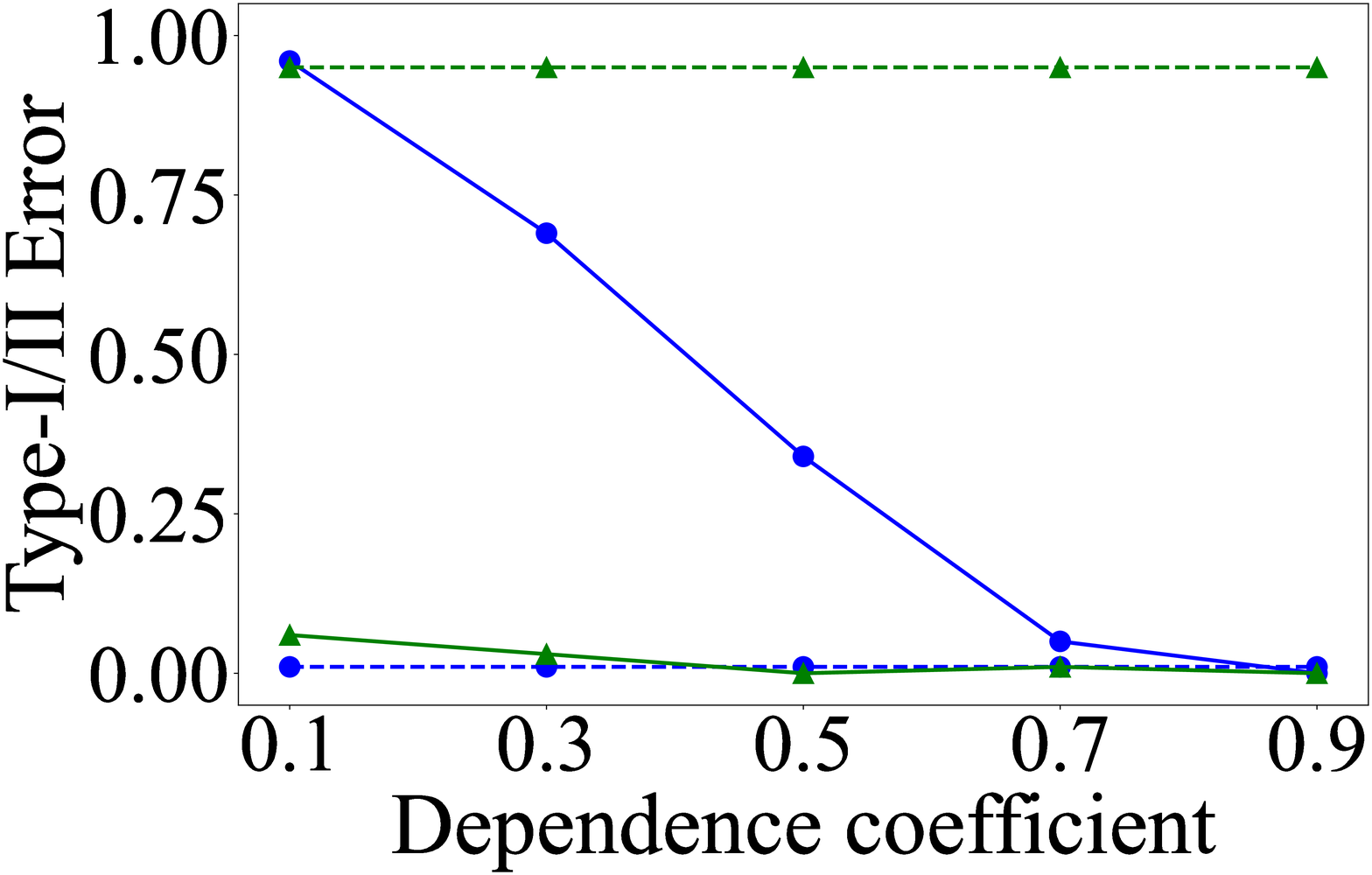}}\\
    
    \subfloat[ER: Case 1 ]{\label{sfig:vnv_b}\includegraphics[width=0.28\textwidth]{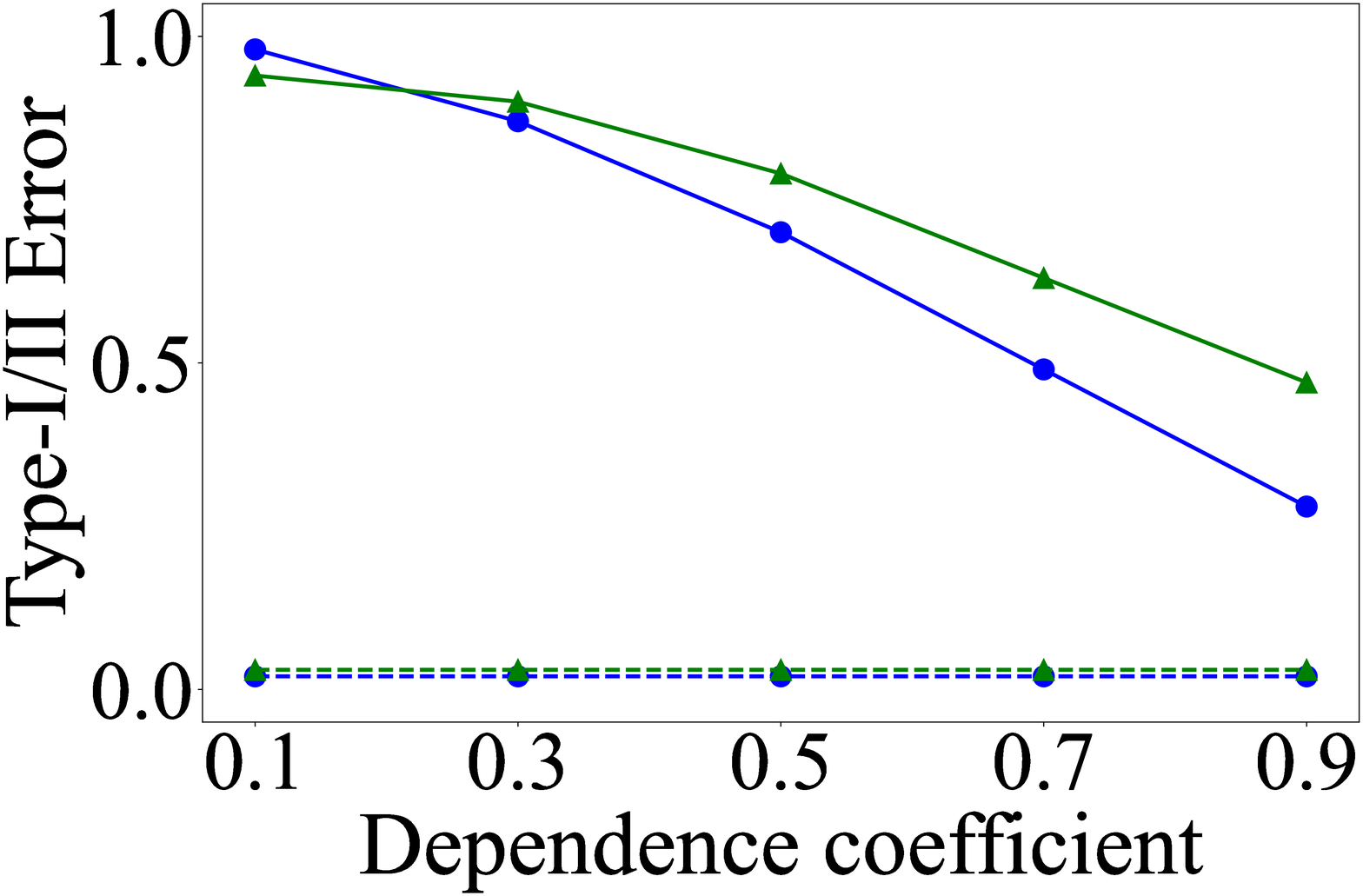}}
    \hspace{1em}
    \subfloat[ER: Case 2 ]{\label{sfig:vnv_d}\includegraphics[width=0.28\textwidth]{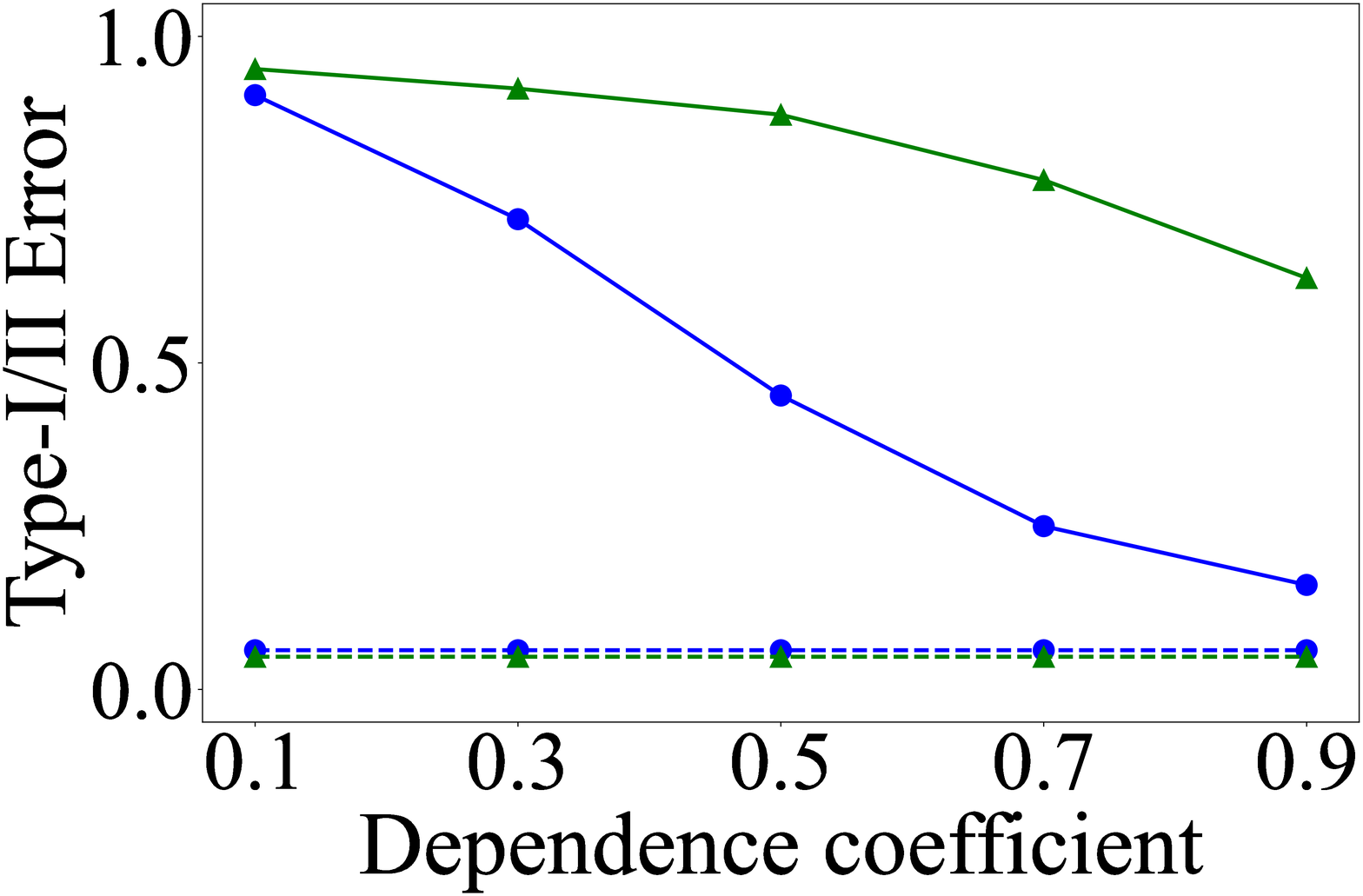}}
    \hspace{1em}
    \subfloat[ER: Case 3 ]{\label{sfig:vnv_f}\includegraphics[width=0.28\textwidth]{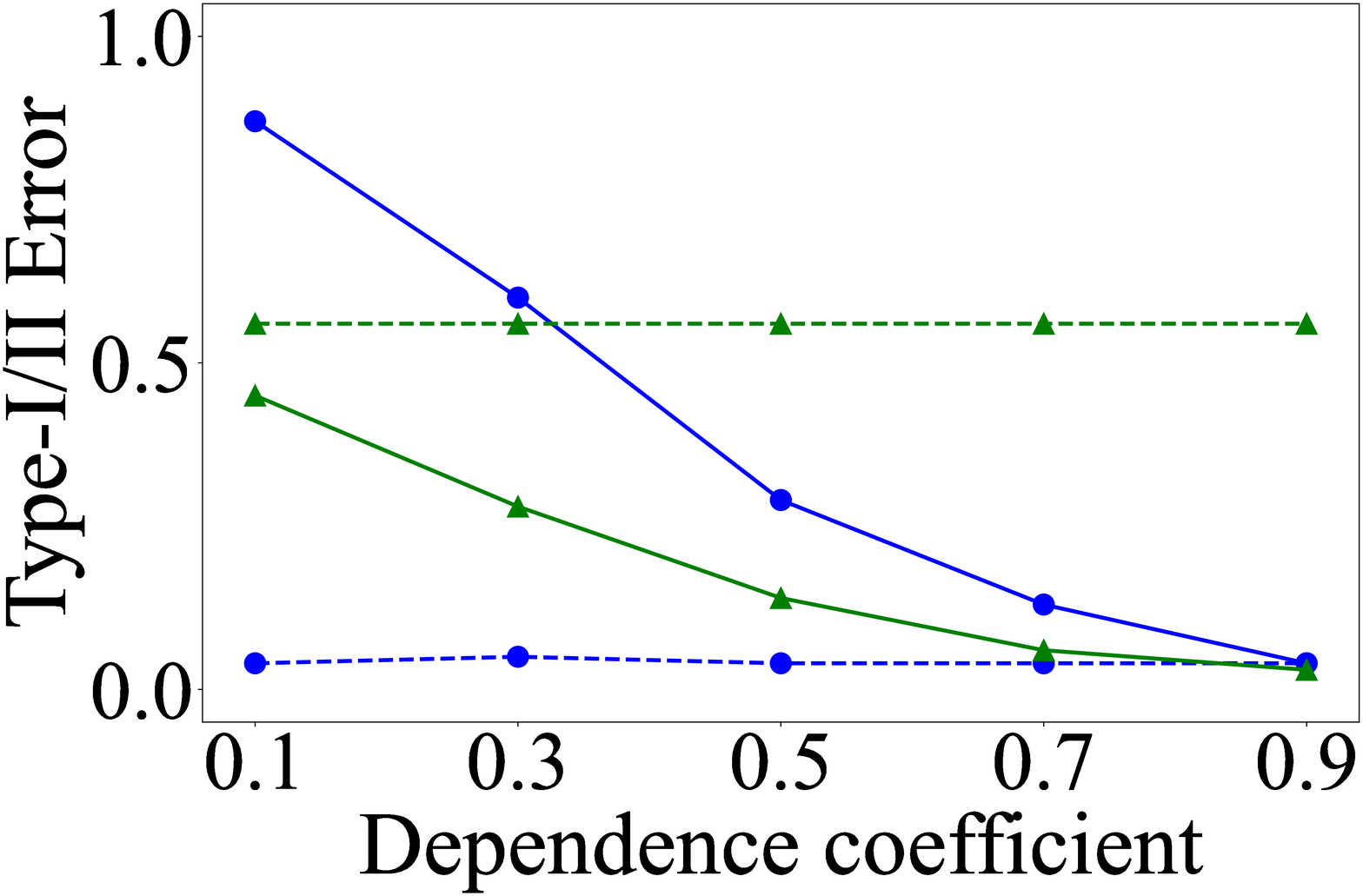}}\\

    \caption{Relational dependence impact on Type I/II errors while variance of noise varied  $\thicksim \mathcal{N}(1, 0.2)$ over multiple trials.}%
    
    \label{fig:dep_vnv}
\end{figure*}

\subsection{Impact of varied noise variance}
\label{ssec:noise_vary}

We conducted an experiment where we draw noise variance from a normal distribution $\sigma^2 \thicksim N(1, 0.2)$ over different trials. From figure \ref{fig:dep_vnv} we can see a slight change of type-II errors compared to Figure 1 in the main paper. However, the trend seems to be very similar.

\subsection{Impact of activation probability on diffusion}

\begin{figure}[!ht]
    \centering
    \label{sfig:ltm_tw}\includegraphics[width=.30\textwidth]{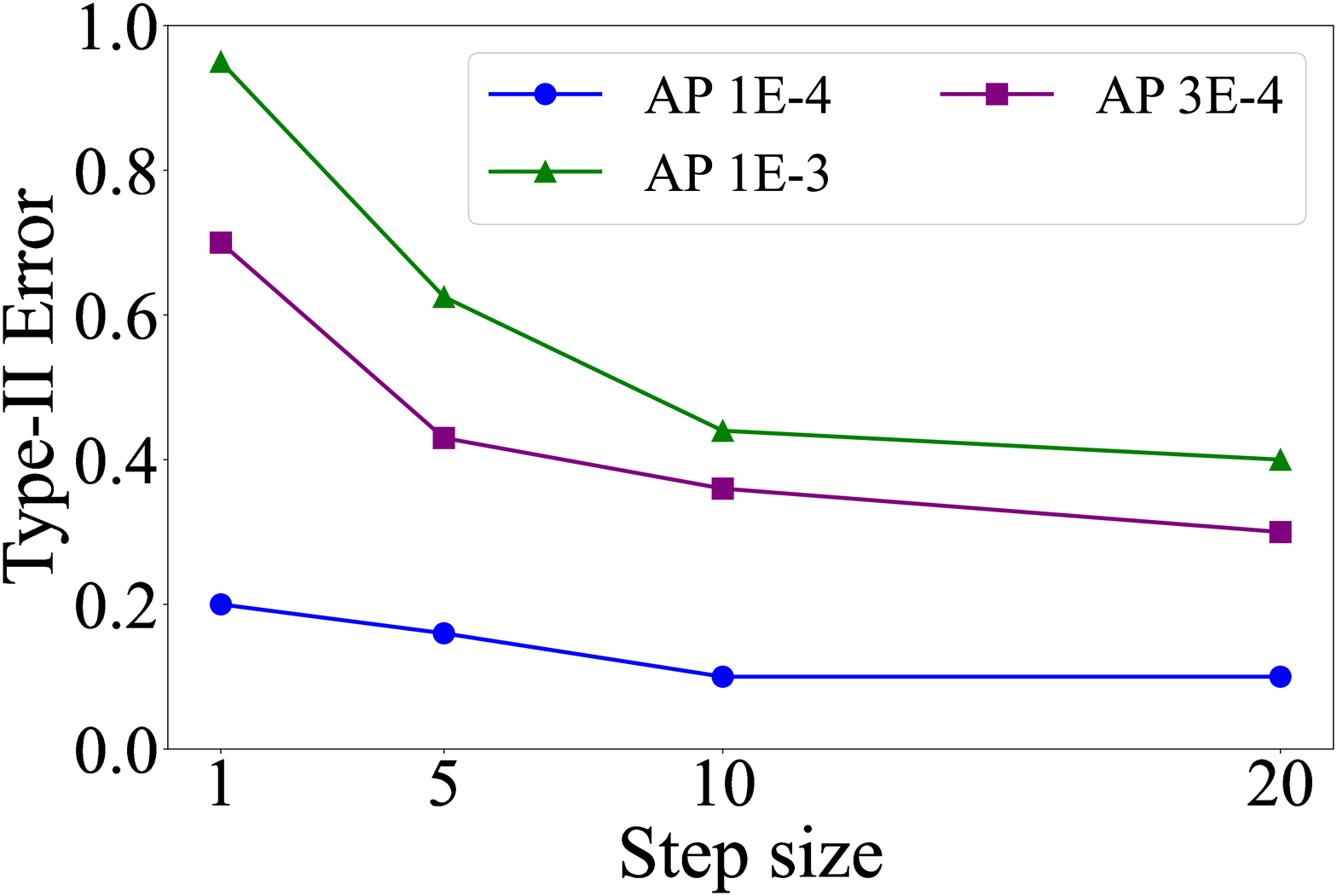}
    
    \caption{Type II error for the Linear Threshold Model on Twitter ego-network.}
    \label{fig:twitter}
\end{figure}

In order to showcase the applicability of the proposed method on large scale real world relational data, we show an extended version of the diffusion experiment from the main paper. We consider a similar semi-synthetic setup with Twitter ego-network which is a larger real world network consisting 11,176 nodes and 1,44,653 edges~\citep{leskovec-nips12}. We consider a sample of 10,000 nodes and vary the initial activation probabilities. Figure \ref{fig:twitter} shows the Type-II errors (y-axis) for different diffusion step sizes (x-axis). The lines correspond to the initial activation probabilities (AP) for the diffusion process. We see the general trend of decreasing Type-II error with higher step sizes. It seems to be almost saturated with step 10. Moreover, the result indicates that the test is sensitive to activation probabilities and with higher activation probability, it shows higher type II error.

\subsection{Comparison to Sobolev Independence Criterion (SIC)}

\begin{figure}[h]
    \centering
    \subfloat{\includegraphics[width=.86\textwidth]{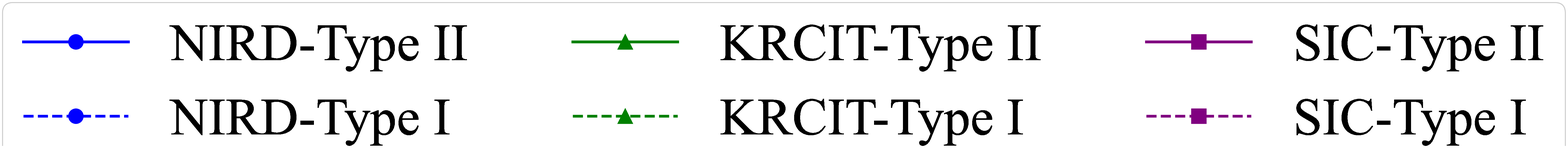}}\\
    
    \setcounter{subfigure}{0}
    \subfloat[Case 1: \text{\ba} ]{\label{sfig:dep_s_a_sic}\includegraphics[width=0.30\textwidth]{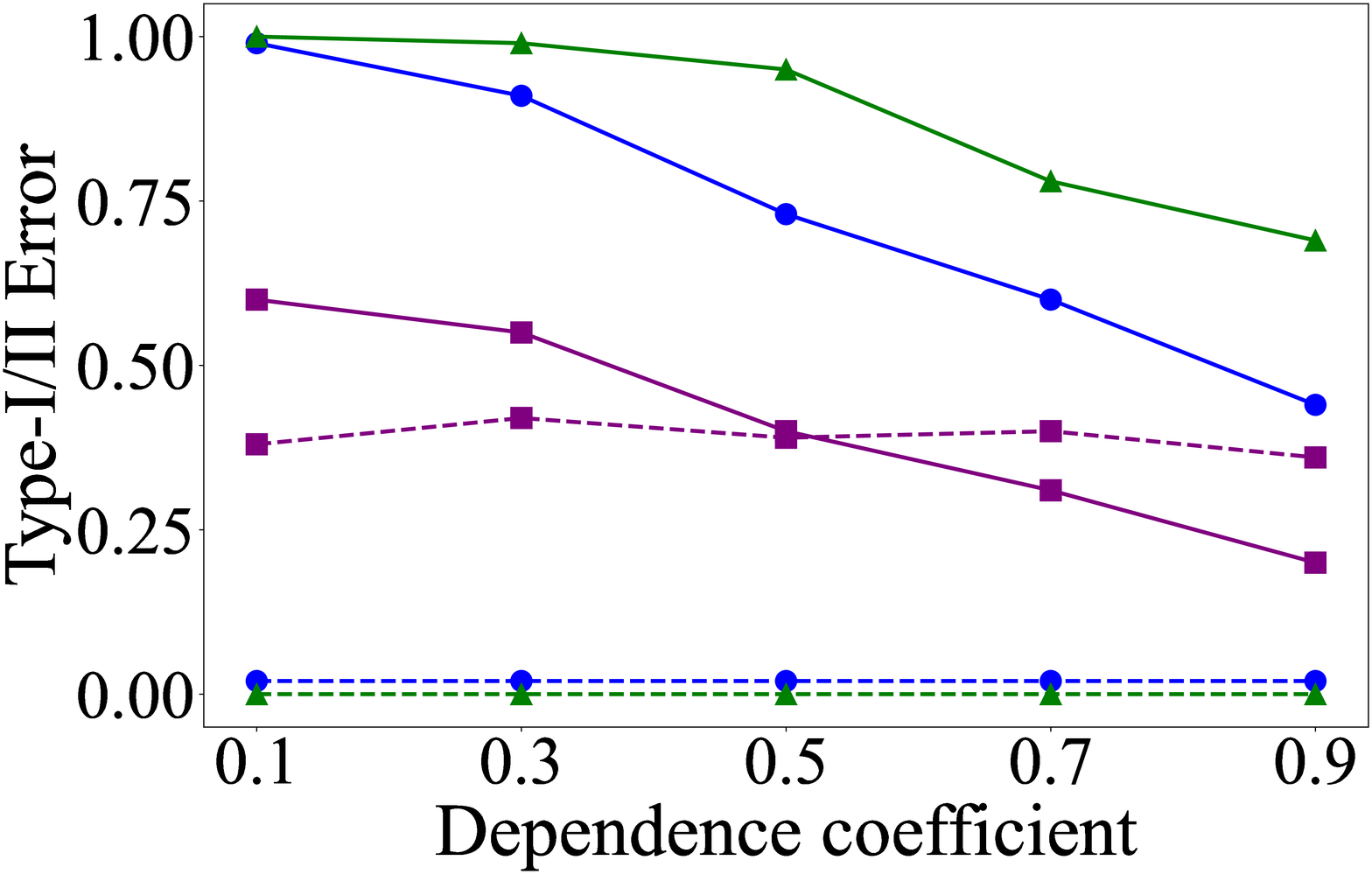}}\hfill
    \subfloat[Case 2: \text{\ba} ]{\label{sfig:dep_s_c_sic}\includegraphics[width=0.30\textwidth]{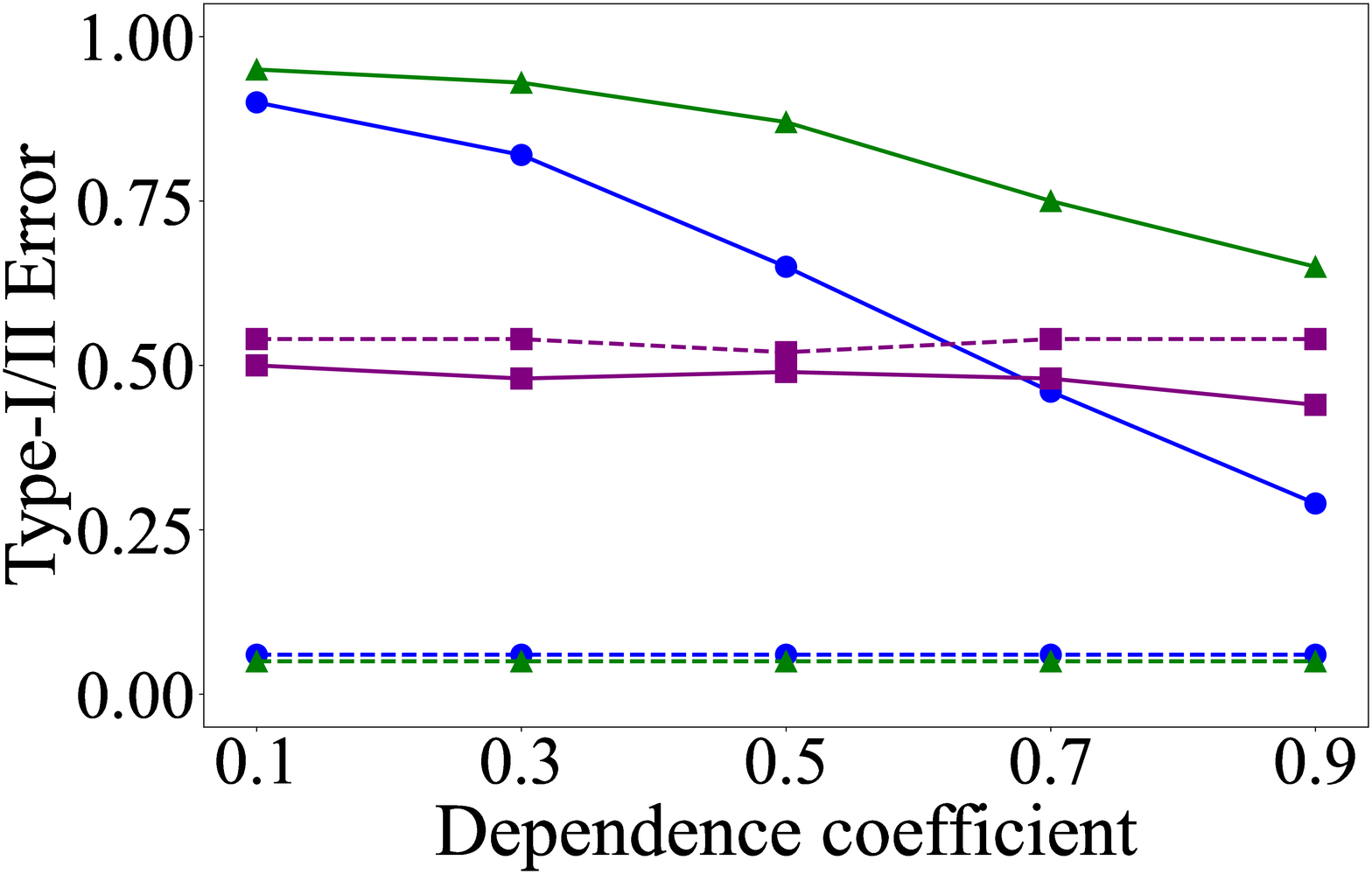}}\hfill
    \subfloat[Case 3: \text{\ba} ]{\label{sfig:dep_s_e_sic}\includegraphics[width=0.30\textwidth]{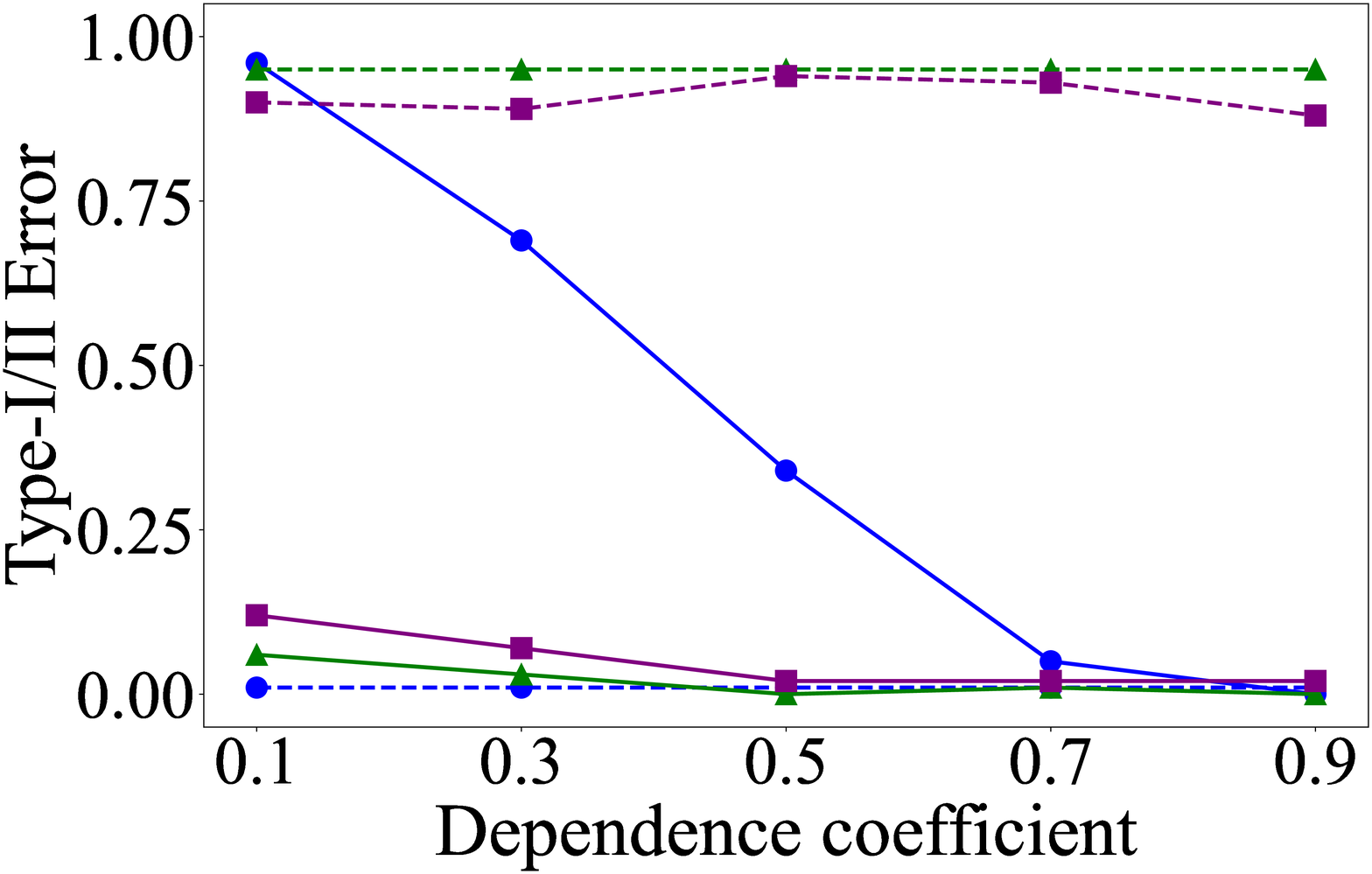}}\\
    
    \subfloat[Case 1: \text{\er} ]{\label{sfig:dep_s_b_sic}\includegraphics[width=0.30\textwidth]{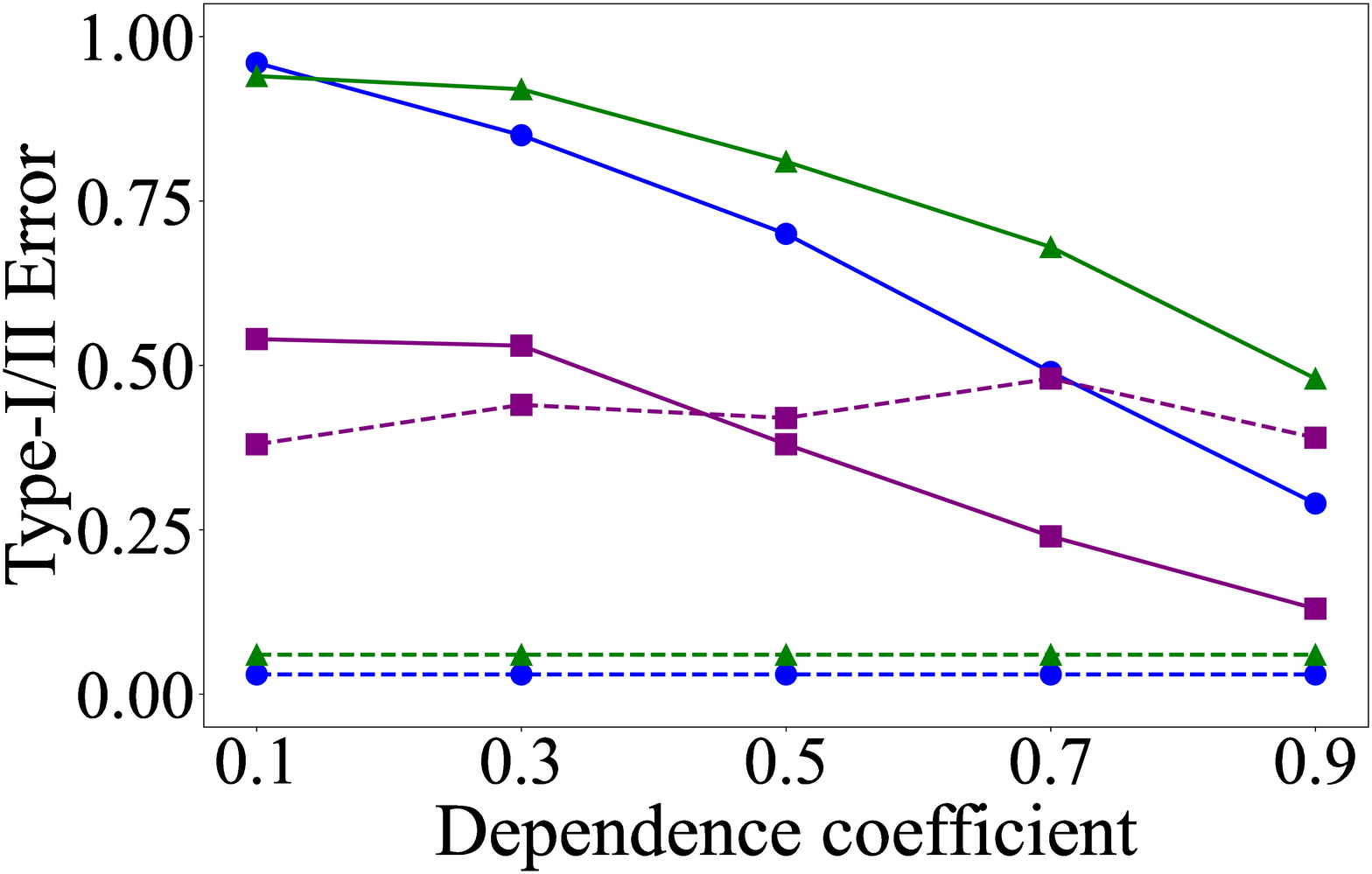}}\hfill
    \subfloat[Case 2: \text{\er} ]{\label{sfig:dep_s_d_sic}\includegraphics[width=0.30\textwidth]{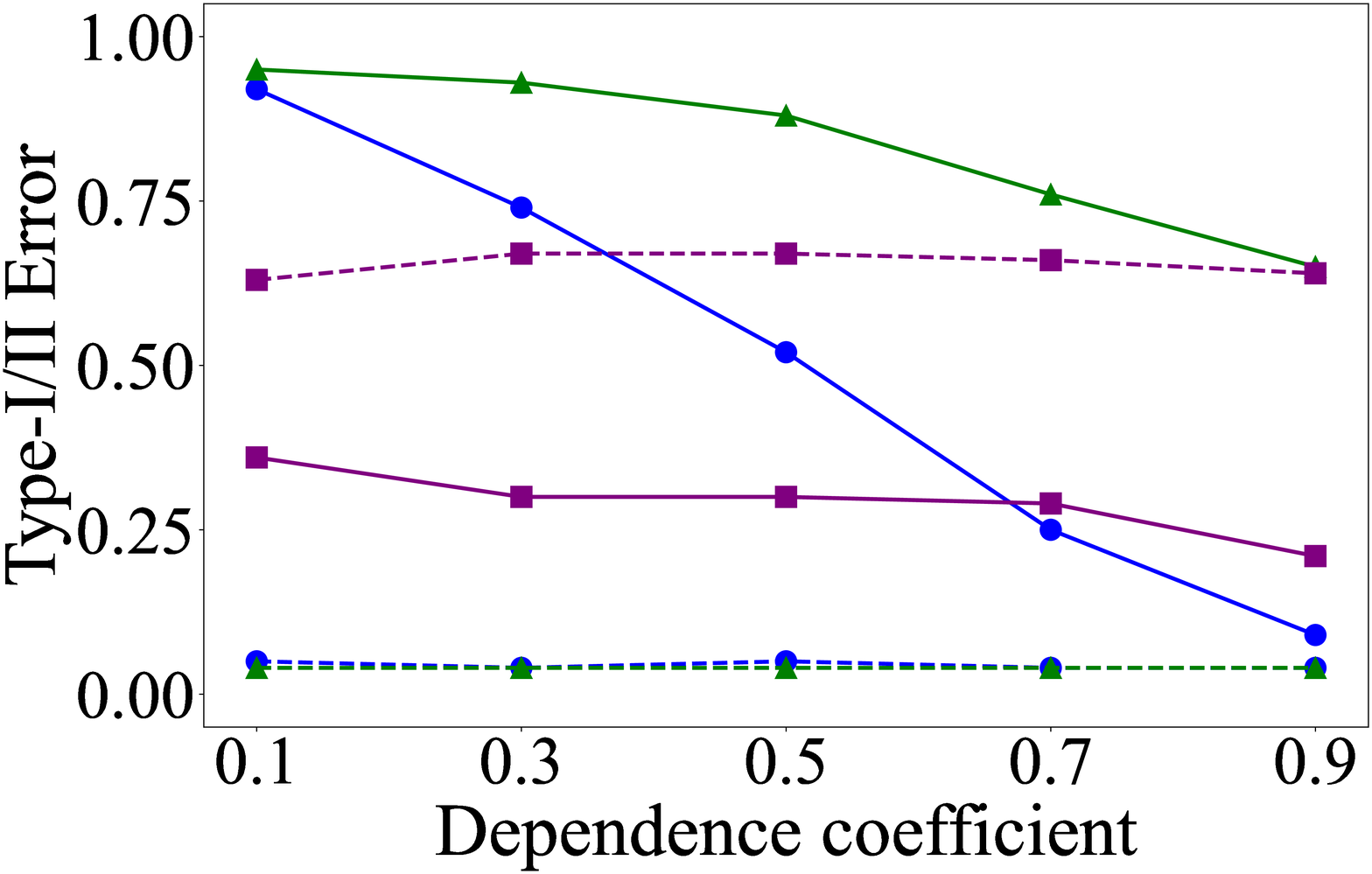}}\hfill
    \subfloat[Case 3: \text{\er} ]{\label{sfig:dep_s_f_sic}\includegraphics[width=0.30\textwidth]{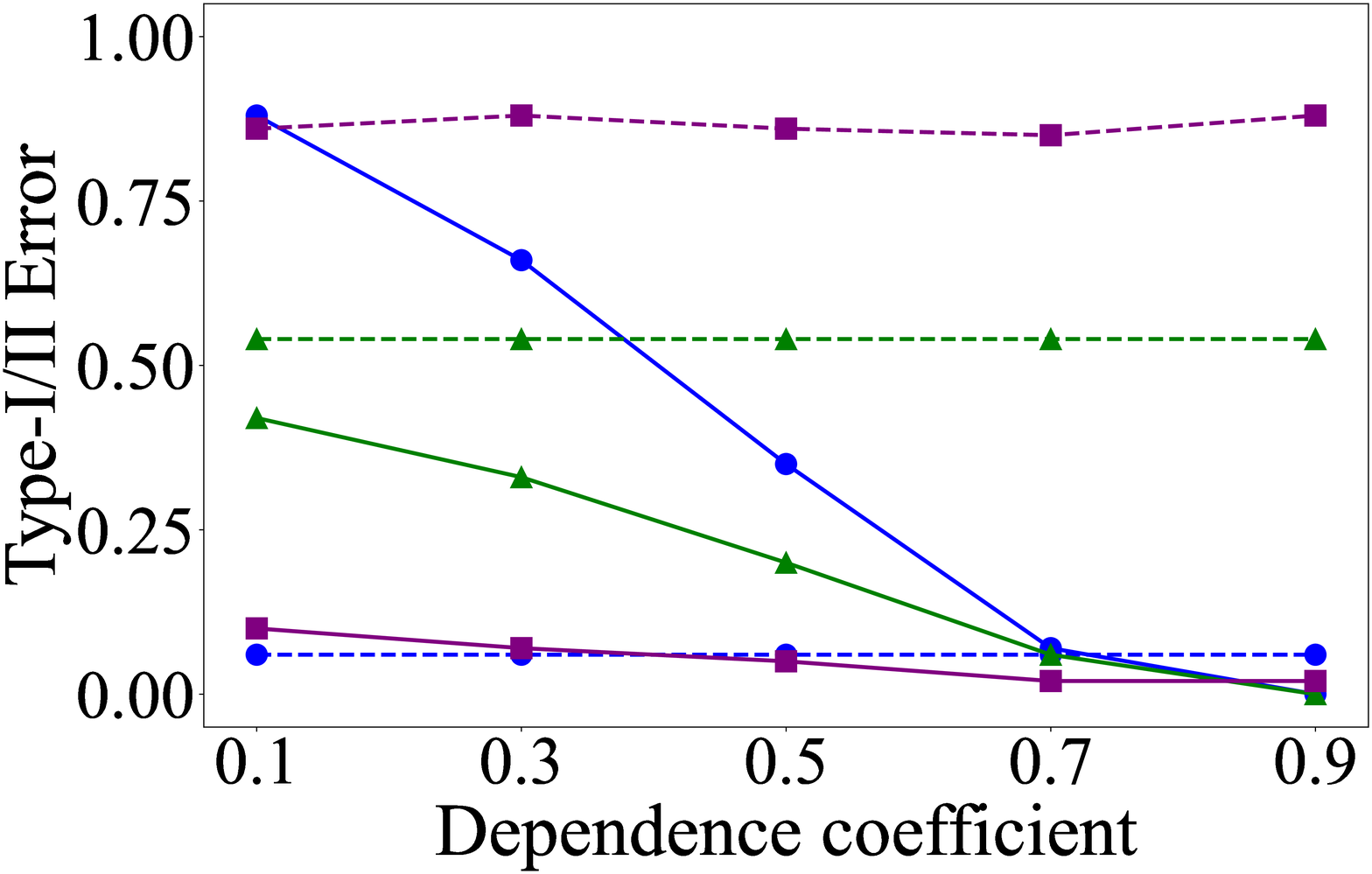}}
    
    \caption{Type I/II errors with polynomial dependency model on synthetic networks for all three cases.}
    
    \label{fig:dep_s_sic}
\end{figure}

To show the effectiveness of relational CI methods vs. CI methods developed for i.i.d. data, we compare both RCI methods (NIRD, KRCIT) to a recent i.i.d. CI test, the Sobolov Independence Criterion~\citep{mroueh-nips19}. SIC is an interpretable dependency measure between multivariate random variables characterized by integral probability metric between the joint distribution and the product of the marginals. We perform the SIC test on the flattened representation of the relational data, similar to KRCIT. Figure \ref{fig:dep_s_sic}  extends the results shown in Figure 1 in the main text. In all three cases, the i.i.d. baseline SIC exhibits high Type I error which shows its poor calibration to reasoning over the relational data. %

\section{Real-world demonstration}
One of the main challenges in social studies is to identify the effect of friends on their peers and the strength of such effects in different domains, e.g. health and violence. Studies show that patterns of interactions among adolescents can reveal possible reasons for changes in their behavior over time. The central question in such studies is how to identify and measure the existence of such effects. The proposed independence test can facilitate reasoning over the existence of dependence between peers' behaviors in social networks by providing a mechanism for falsifying statistical hypotheses.

As a demonstration, we examine the 50 Women dataset~\citep{michell-ssm97}. This dataset has the smoking, sport, drug, alcohol consumption habits of 50 female students, along with their friendship information, over the course of three years. Each of the behavioral variables are coded as categorical variables indicating how regularly women engage in each of the behaviors.
Assuming independence between the behavior peers as the null hypothesis, the goal of  this analysis is to explore whether the habits of a student's friends are associated with her habits in subsequent years. 

\begin{table}
\caption{Real-world demonstration: exploration of the dependence between the habits of students and their first-hop neighbors in 50 Women dataset }
\centering
\small\addtolength{\tabcolsep}{-2pt}
\begin{tabular}{lrrrrrrr}  
period& attribute & attribute type & t0& \mcit\_all & \mcit\_t0\\
\midrule
    1 $\rightarrow$ 2 &   alcohol &      binary &   4 &  0.425532 &  0.000000 \\
    1 $\rightarrow$ 2 &   alcohol &  categorical &   NA &  0.000000 & NA \\
    1 $\rightarrow$ 2 &      drug &      binary &  35 &  0.000000 &  0.138298 \\
    1 $\rightarrow$ 2 &      drug &  categorical &   NA &  0.000000 & NA \\
    1 $\rightarrow$ 2 &     smoke &      binary &  35 &  0.000000 &  0.021277 \\
    1 $\rightarrow$ 2 &     smoke &  categorical &   NA &  0.000000 & NA \\
    1 $\rightarrow$ 2 &     sport &      binary &  12 &  0.925532 &  0.978723 \\
    1 $\rightarrow$ 2 &     sport &  categorical &   NA &  0.925532 & NA \\
    1 $\rightarrow$ 2,3 &   alcohol &      binary &   5 &  0.114583 &  0.197917 \\
    1 $\rightarrow$ 2,3 &   alcohol &  categorical &   NA &  0.000000 & NA \\
    1 $\rightarrow$ 2,3 &      drug &      binary &  35 &  0.000000 &  0.583333 \\
    1 $\rightarrow$ 2,3 &      drug &  categorical &   NA &  0.000000 & NA \\
    1 $\rightarrow$ 2,3 &     smoke &      binary &  36 &  0.000000 &  0.000000 \\
    1 $\rightarrow$ 2,3 &     smoke &  categorical &   NA &  0.000000 & NA \\
    1 $\rightarrow$ 2,3 &     sport &      binary &  12 &  1.000000 &  0.166667 \\
    1 $\rightarrow$ 2,3 &     sport &  categorical &   NA &  1.000000 & NA \\
    2 $\rightarrow$ 3 &   alcohol &      binary &   3 &  0.125000 &  0.666667 \\
    2 $\rightarrow$ 3 &   alcohol &  categorical &   NA &  0.281250 & NA \\
    2 $\rightarrow$ 3 &      drug &      binary &  32 &  0.000000 &  0.125000 \\
    2 $\rightarrow$ 3 &      drug &  categorical &   NA &  0.000000 & NA \\
    2 $\rightarrow$ 3 &     smoke &      binary &  31 &  0.000000 &  0.281250 \\
    2 $\rightarrow$ 3 &     smoke &  categorical &   NA &  0.000000 & NA \\
    2 $\rightarrow$ 3 &     sport &      binary &  20 &  0.864583 &  0.479167 \\
    2 $\rightarrow$ 3 &     sport &  categorical &   NA &  0.864583 & NA \\
\bottomrule

\end{tabular}

\label{OurMethodVSBaseline}
\end{table}
Table \ref{OurMethodVSBaseline} shows p-values estimated by our kernel test method considering four attributes in 50 Women dataset. We use column $Period$ to indicate the years we consider for the test, e.g., in $Period$ $1 \rightarrow 2,3$, we explore students' behavior change from first year to the second and third year. 
We consider both the original categorical coding and a binarization of the categorical attributes, which is 1 if the student uses a substance at least once during the year and 0 otherwise.
The number of students who did not engage in the behavior during the first time point is shown in column $t0$, e.g, in the first row of the table, 4 students did not drink alcohol in the first year. We exclude $t0$ for categorical data (indicated by NA) because the frequency of the habit is intrinsic to the hypothesis of interest in these cases. The last two columns ( $NIDR\_all$ and $NIDR\_t0$ ) show p-values measured by NIRD. In $NIDR\_all$ and $NIDR\_t0$
we consider all women (whether they have the habit in a year or not), and women who do not have the habit in the first time point, respectively. 
Overall we find:
\begin{itemize}[leftmargin=*]
    \item Sports activity of peers is not associated with whether a student plays a sport or not. High values of $NIDR\_t0$ and $NIDR\_all$ are enough evidences to accept the null hypothesis of independence.
    \item Peer smoking habits are associated with students' frequency of smoking: $NIDR\_all = 0 $ and $NIDR\_t0 < 0.022 $ for all time periods, except period $2 \rightarrow 3$ where $NIDR\_t0 \approx 0.28$.
    \item Peer drug use is not associated with subsequent drug use in previously non-drug using students ( $NIDR\_t0 > 0.05$). However, when we consider the effect of drug users on non-drug users and vice versa, it becomes associated in both the use and rate of consumption ($NIDR\_all =0$). 
    \item Peer alcohol consumption is associated with the level subsequent alcohol use ($NIDR\_all$ $ =0$, except in period $2 \rightarrow 3$ where $NIDR\_all > 0.1$ ),
    but not with the decision for a non-drinking student tot begin drinking.
\end{itemize}

Different studies \citep{michell-ssm97,pearson-depp00} deploy 50 women data to explore the association between gender, risk-taking or social position and smoking or drug usage in groups of youngsters. In particular our results comport with Pearson et al. \citep{pearson-depp00} who show that drug usage and smoking are contagious among group of friends who are highly connected and people who are loosely connected to a friendship group.

\end{document}